\documentclass[]{bytedance_seed}
\usepackage[utf8]{inputenc}
\usepackage{amsmath,amssymb,array,tabularx,float,longtable,url}
\usepackage{placeins}
\newcolumntype{Y}{>{\raggedright\arraybackslash}X}
\newcommand{\bench}{\textsc{TabularMath}}
\graphicspath{{figures/}}

\newcommand{\JointHighFit}{42}
\newcommand{\JointPrecisionGap}{33}

\newcommand{\BestBaselineName}{PySR}
\newcommand{\BestBaselineShift}{76.1}
\newcommand{\BestBaselineSolved}{72}

\newcommand{\FlatMedianRtwo}{-1.67}

\newsavebox{\fitbox}
\newcommand{\fitwidth}[1]{\sbox{\fitbox}{#1}\ifdim\wd\fitbox>\linewidth\resizebox{\linewidth}{!}{\usebox{\fitbox}}\else\usebox{\fitbox}\fi}
\providecommand{\bibstylename}{plainnat}

\renewcommand{\bibstylename}{plainnat}

\title{TabularMath: Evaluating Computational Extrapolation in \\ Tabular Learning via Program-Verified Synthesis}

\author[1,2,*,\dagger]{Zerui Cheng}
\author[1,\dagger]{Jiashuo Liu}
\author[2]{Jianzhu Yao}
\author[2]{Pramod Viswanath}
\author[1]{Ge Zhang}
\author[1]{Wenhao Huang}
\renewcommand\authorlist{%
  \authorformat[1,2,*,\dagger]{Zerui Cheng}, \authorformat[1,\dagger]{Jiashuo Liu}, \authorformat[2]{Jianzhu Yao},\\[3pt]
  \authorformat[2]{Pramod Viswanath}, \authorformat[1]{Ge Zhang}, \authorformat[1]{Wenhao Huang}}

\affiliation[1]{ByteDance Seed}
\affiliation[2]{Princeton University}

\contribution[*]{Work done at ByteDance Seed}
\contribution[\dagger]{Corresponding authors}

\abstract{Standard tabular benchmarks reward interpolation: predicting values inside the data manifold, where local statistical smoothing works. However, much high-value tabular data, from pricing and financial rules to physical simulation, is produced by deterministic computations rather than noisy relationships. Can tabular models move from statistical interpolation to \emph{computational extrapolation}, returning exact answers even beyond the range seen in training? We introduce \bench{}, a benchmark of 114 deterministic computations (233,472 rows) compiled from GSM8K and AIME problems into verified programs. We evaluate nine tabular methods, closed-form and symbolic regression baselines, and in-context learning with GPT-OSS-120B. On regression metrics, TabPFN v2.5 looks excellent: with 128 rows its median $R^2$ is 0.999 in-distribution, and it is the only tabular method whose median $R^2$ stays positive (0.76) when nearly every answer exceeds the training range. Exact answers tell a different story. Out of distribution (OOD), its accuracy falls from 39.5\% to 7.8\%, it answers no query correctly in 70 of 100 GSM8K tasks, and sixteen times more rows raise its OOD accuracy by only 0.4 points. Yet the answers are in the table: by construction, every label is a program's output on the row's inputs, and the training rows typically suffice to recover that program. Symbolic regression on the numbers alone recovers every OOD answer in 72 of 100 GSM8K tasks, and in-context learning with GPT-OSS-120B answers 39.2\% of OOD queries. Tabular foundation models fit the trend but miss the computation, an inductive-bias gap that more data does not close. The competition-level AIME tasks remain open for every method.
}

\date{\today}

\correspondence{\begin{tabular}[t]{@{}l@{\ }l@{\ }l@{}}Zerui Cheng & at & \email{zerui.cheng@princeton.edu}\\Jiashuo Liu & at & \email{liujiashuo.77@bytedance.com}\end{tabular}}

\begin{document}
\maketitle
\section{Introduction}
\begin{figure}[t]
\centering
\includegraphics[width=\linewidth]{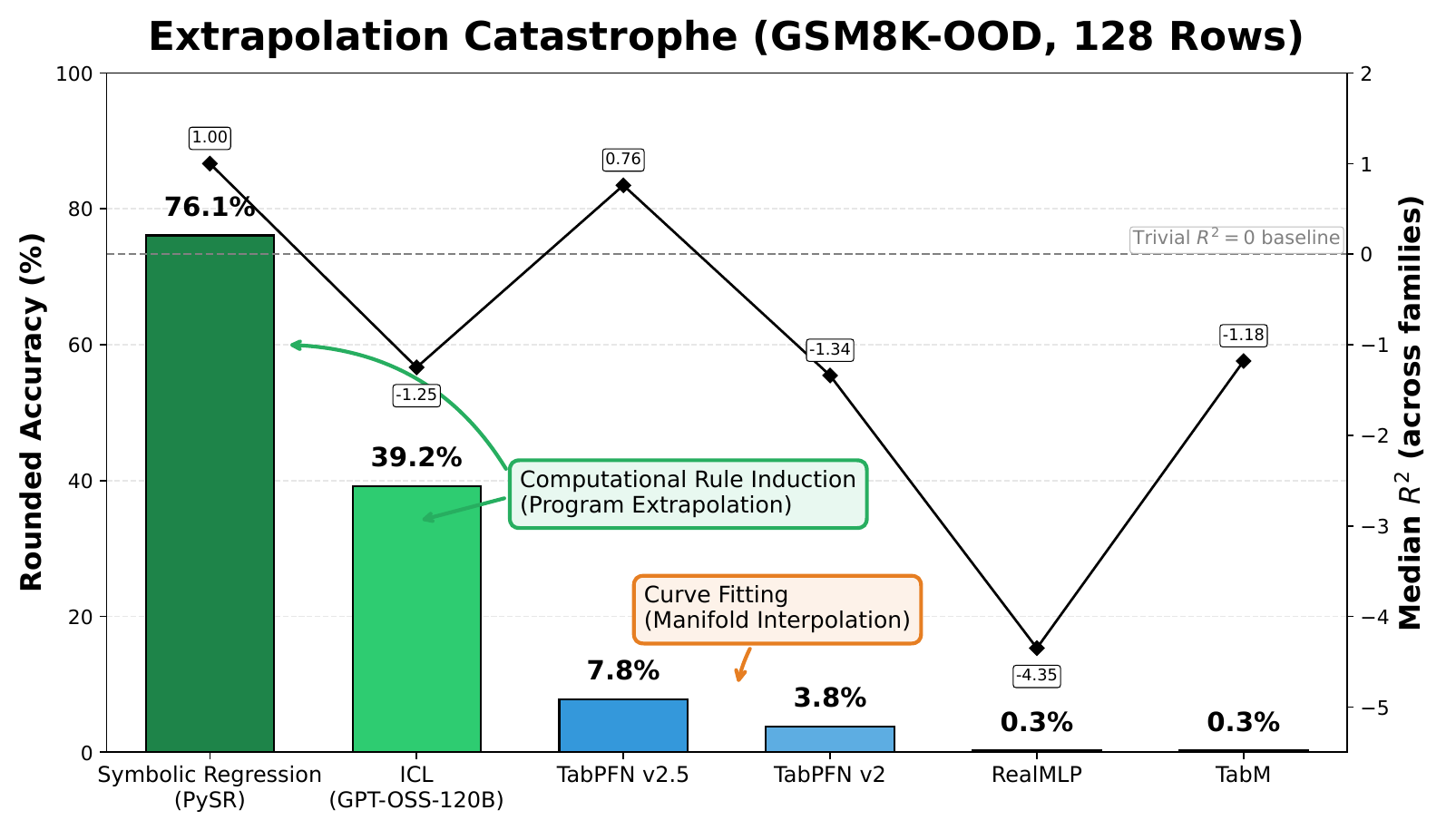}
\caption{\emph{The $R^2$--accuracy gap.} GSM8K at a 128-row cap on the out-of-distribution (OOD) split, where nearly every query's answer lies above all training answers. Bars show rounded accuracy (left axis); diamonds show median $R^2$ across families (right axis). Methods that induce the computation return exact answers: symbolic regression (PySR) answers 76.1\% of OOD queries and in-context learning (ICL) with GPT-OSS-120B 39.2\%, while every tabular method stays below 10\%. TabPFN v2.5 nonetheless reaches a median $R^2$ of 0.76, the only positive median among the tabular methods: it captures the functional trend while missing the exact answers. Values are from Table~\ref{tab:baselines}.}
\label{fig:gap}
\end{figure}

Can modern tabular models learn to \emph{compute}? Given a table of input--output pairs generated by a deterministic rule, can a model recover the rule and return exact answers on inputs outside the training range?

Tabular benchmarks mainly evaluate interpolation: predicting values inside a data manifold, where models that smooth locally are rewarded. Many high-value tables, however, are generated by deterministic computations rather than noisy relationships. A table of quantities, prices, and tax rates may record the outputs of a fixed pricing rule, and tables from financial rules or physical simulation arise in the same way. To reproduce such a rule, a predictor must move from statistical interpolation to \emph{computational extrapolation}: it must return the exact answer---here, the target rounded to the nearest integer---including for rows whose answers are larger than any seen in training. A high $R^2$ does not guarantee this, because predictions can follow the trend of the targets while missing every exact answer (Figure~\ref{fig:gap}).

Consider a one-unit error. Suppose integer targets have variance $\sigma_y^2$, and a predictor returns $\hat y=y+1$ for every query row. Its rounded accuracy is zero, while $R^2=1-1/\sigma_y^2$. As the target variance grows, the regression score approaches one without a single correct rounded answer. Do modern tabular predictors behave this way when learning mathematical computations?

Recent tabular foundation models make this question timely. Prior-fitted networks predict well from small tables \citep{hollmann2023tabpfn,hollmann2025tabpfn,tabpfn25}, as do tree ensembles and recent neural architectures \citep{grinsztajn2022tree,gorishniy2024tabm,mcelfresh2023neural}. Arithmetic networks and symbolic regression explicitly seek numerical rules that extend beyond training examples \citep{trask2018nalu,madsen2020nau,cranmer2023pysr}. \bench{} brings these lines of work together: its deterministic computations, drawn from human-written math problems, test whether tabular predictors extend a numerical relationship with answer-level precision.

\bench{} starts from mathematical problems in GSM8K \citep{cobbe2021training} and AIME. Each original problem, or \emph{seed}, becomes a \emph{task family} (or simply \emph{family}): a computation with variable inputs and a program that supplies its answers. The inputs become columns, and running the program labels 2,048 rows for each of the 114 families. The \emph{out-of-distribution (OOD) split} sorts rows by output and holds out the upper 20\% as queries: nearly every query's answer lies above all training answers, while most queries keep every input within its training range. A \emph{random split} of the same rows provides the in-distribution comparison.

The results are stark. TabPFN v2.5, the strongest tabular method, fits the numerical relationship well but rarely returns the correct rounded answer on the OOD split: its accuracy falls from 39.5\% on the random split to 7.8\%, \JointPrecisionGap{} families have $R^2>0.8$ and accuracy below 20\%, and more training rows raise its random-split accuracy but not its OOD accuracy.

Yet the answers are in the table. By construction, every label is the output of a program on the row's inputs, so nothing beyond the numbers is needed to compute it. For most families, the rows a learner sees are also enough to recover the rule: symbolic regression with PySR, which uses only the numbers, \emph{solves} 72 of 100 GSM8K families from the same 128-row samples, answering every OOD query exactly, while TabPFN v2.5 solves 2. The failure of tabular foundation models is therefore one of inductive bias, not of information. In-context learning with GPT-OSS-120B, given the same small tables, answers 39.2\% of OOD queries, against 8.9\% for TabPFN v2.5 on the same families (Section~\ref{sec:llm}); it solves about a third of these families and answers no OOD query in most of the others.

Our main contributions are:
\begin{enumerate}
\item \textbf{A general framework for synthesizing verified tabular benchmarks.} Coupled generator--verifier programs turn a mathematical problem into a table in which an executed program computes every label, and the split controls which part of the output range is queried.
\item \textbf{\bench{}, a benchmark for computational extrapolation in tabular learning.} It comprises 114 computations from GSM8K and AIME and 233,472 rows; each family is a separate regression task with exact labels.
\item \textbf{A study of interpolation versus extrapolation.} Evaluating nine tabular methods, closed-form and symbolic baselines, and in-context learning with GPT-OSS-120B across row caps, we show that high $R^2$ and correct answers come apart, that more data does not close the gap, and that a numbers-only symbolic method recovers most answers from the same rows.
\item \textbf{Insights for reasoning-capable tabular models.} Per-family analysis identifies four distinct behaviors of the evaluated methods, and classifying the families by computation type shows which computations a short closed-form fit recovers but tabular foundation models miss. Together these point to program induction, arithmetic modules, and hybrid predictors.
\end{enumerate}

\section{Executable Tabular Task Families}
\label{sec:construction}
\begin{figure}[t]
\centering
\includegraphics[width=0.78\linewidth]{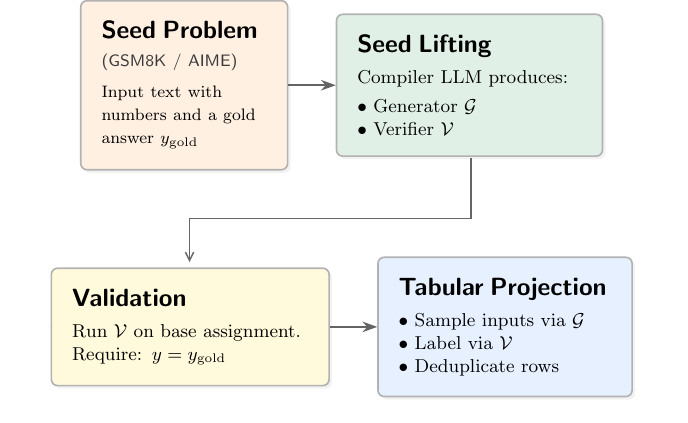}
\caption{\textbf{From seed problem to labeled table.} Seed lifting compiles a seed into a generator and a verifier, validation anchors the verifier to the seed answer, and tabular projection samples inputs with the generator and labels them with the verifier.}
\label{fig:construction}
\end{figure}
\paragraph{From a mathematical problem to a table.}
We build each of the 114 \bench{} families from a seed, a natural-language problem with its answer. GPT-5 with high reasoning effort \citep{openai2025gpt5}, acting as a compiler, identifies the variable quantities (the \emph{slots}) and writes two programs: a generator $G$ that samples valid inputs $x$, and a verifier $V$ that checks each input and computes its target $y$ (Figure~\ref{fig:construction}). Each family produces the dataset
\begin{equation}
\mathcal D_i=\{(x_{ij},V_i(x_{ij}))\}_{j=1}^{2048},\qquad x_{ij}\sim G_i.
\end{equation}
Each family is a separate regression task, and the learner sees only its table.

\begin{figure}[t]
\centering
\includegraphics[width=0.85\linewidth]{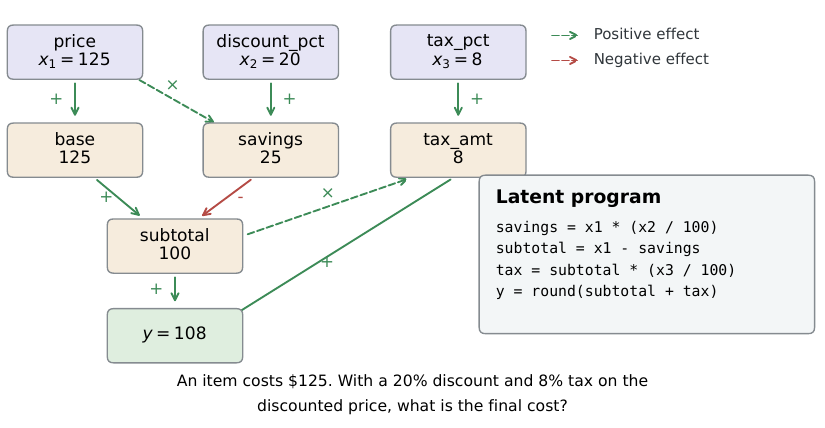}
\caption{\textbf{The target is an executed computation.} The pricing example as a graph of intermediate quantities and as a short program. Green edges increase and red edges decrease the downstream quantity; edges marked $\times$ multiply. A 20\% discount on 125 saves 25, which is subtracted to give a subtotal of 100, and 8\% tax on the subtotal gives a total of 108. Learners observe only input--output rows; the program, which plays the role of the verifier, determines the labels.}
\label{fig:program}
\end{figure}

\paragraph{A pricing example.}
Figure~\ref{fig:program} traces an illustrative pricing rule. It maps a base price $p$, discount $d$, and tax rate $t$ to the target $p(1-d)(1+t)$: an item costing \$125 with a 20\% discount and 8\% tax costs \$108. Each row records the rule's inputs and the computed total, for example $(\$200, 15\%, 10\%)\to\$187$ or $(\$90, 25\%, 5\%)\to\$71$. Larger prices or tax rates produce totals beyond the training range, while the calculation stays fixed. The benchmark also includes multi-step and competition-mathematics computations with less direct dependencies. Appendix~\ref{app:worked} follows another illustrative family from seed problem to labeled rows, and Appendix~\ref{app:aime-example} works through an AIME family from the benchmark.

\paragraph{Seeds and validation.}
The 100 GSM8K seeds were drawn uniformly at random from the GSM8K test split, and the 14 AIME seeds are the AIME 2024 problems whose families yield at least 2,048 distinct rows during construction (Table~\ref{tab:scale}). A family is accepted only if its verifier reproduces the seed answer, respects type and range constraints, and finishes within the time limit; schema, generation, and execution failures are fed back to the compiler, with at most 20 synthesis attempts. These checks anchor the labels to the seed, and the verifier then labels freshly sampled inputs. Each accepted family is stored as a specification: the seed's input values, the input constraints, and the verifier. Rows are deduplicated during construction (Appendix~\ref{app:records}).

\begin{table}[t]
\centering
\caption{\textbf{The benchmark comprises 114 families of 2,048 rows each.} Feature counts include the derived features supplied to the predictors.}
\label{tab:scale}
\begin{tabular}{lrrrr}
\toprule
Source & Families & Rows & Mean features & Feature range\\
\midrule
GSM8K & 100 & 204,800 & 45.3 & 25--98\\
AIME 2024 & 14 & 28,672 & 40.6 & 18--88\\
\bottomrule
\end{tabular}
\end{table}

\paragraph{What the learner observes.}
Whether a family uses arithmetic, conditionals, iteration, integer constraints or search, it exposes the same tabular interface: input columns and a target, with no intermediate steps. The numeric slots, which we call the \emph{primary slots}, form the main columns. Derived features add log-magnitude, sign, parity, modular and categorical features, plus the greatest common divisor of the integer slots and the number of slot pairs in which one divides the other; they expose numerical structure without supplying the computation that combines the quantities into an answer. Every tabular method receives the same columns, with feature counts in Table~\ref{tab:scale}. Predictors may interpolate, infer a formula, or use another representation; evaluation concerns only their outputs. Because labels are noise-free, every error is attributable to the predictor and traceable to a specific row.

\section{Evaluation Design}
\label{sec:evaluation}
\paragraph{Row caps and splits.}
We evaluate each family at a row cap $r\in\{32,64,128,256,512,1024,2048\}$: we subsample $r$ rows and split them 80/20 into training rows and queries, so that cap 128 gives 102 training rows and 26 queries. The random split samples both sets at random from the subsample. The OOD split sorts the subsample by $y$ and holds out its upper 20\%, so it tests prediction in the upper tail of the outputs: at cap 128, 96.2\% of OOD GSM8K queries have a target above every training target. Sorting by output also shifts the inputs that raise $y$, but most queries stay in range: 82.3\% keep every primary slot inside its training range (Table~\ref{tab:inrange}, Appendix~\ref{app:inrange}). The OOD split therefore tests extrapolation in the output.

\paragraph{Two complementary scores.}
For family $i$ with query rows $Q_i$, \emph{rounded accuracy} is
\begin{equation}
 A_i=\frac{100}{|Q_i|}\sum_{j\in Q_i}
 \mathbf 1\{\operatorname{round}(\hat y_{ij})=\operatorname{round}(y_{ij})\}.
\end{equation}
In 99 of the 114 families every target is an integer; rounded accuracy applies the same unit-width resolution to the others. We average $A_i$ equally across families and report the median family-level $R^2$ to describe the typical family; unless noted, ``accuracy'' means rounded accuracy, and ``OOD accuracy'' its value on the OOD split. $R^2$ is computed on each family's query rows against the query-set mean, so it is negative when the predictions do worse than that constant. A method \emph{solves} a family on a split if it answers every query exactly. For the closed-form and symbolic baselines we also report tolerance accuracy, the share of queries within 0.5\%, 1\% or 5\% relative error (Appendix~\ref{app:tolerance}).

The two scores measure different properties. $R^2$ is scale-free, while rounding fixes an absolute answer resolution. An $R^2$ of 0.99 still allows a root-mean-square error of a tenth of the target's standard deviation, which is more than half a unit once that deviation exceeds 5. $R^2$ alone therefore cannot tell whether a learner has found the computation; rounded accuracy tests this directly, and $R^2$ shows how closely the predictions follow the targets when they miss.

\paragraph{Numerical-resolution check.}
Rounded scoring requires stored targets that keep integer resolution. Float32 conversion preserves the rounded target for every GSM8K row. One AIME family, 2024-I-14, contains targets as large as $2.51\times10^{10}$, and float32 conversion changes the rounded target on 610 of its 2,048 rows. The primary AIME analysis therefore uses the 13 \emph{numerical-resolution-stable} families, whose rounded targets survive that conversion unchanged.

\paragraph{Tabular methods.}
The nine tabular methods are XGBoost \citep{chen2016xgboost}, CatBoost \citep{prokhorenkova2018catboost}, LightGBM \citep{ke2017lightgbm}, Random Forest \citep{breiman2001random}, TabPFN v2 and v2.5 \citep{hollmann2025tabpfn,tabpfn25}, RealMLP-TD-S \citep{holzmuller2025realmlp}, TabM \citep{gorishniy2024tabm}, and xRFM \citep{beaglehole2025xrfm}. They use fixed configurations; TabPFN v2.5 uses \texttt{v2.5\_default} with eight estimators. Input and target standardization use training-split statistics. The four tree ensembles predict piecewise-constant functions and cannot output a value above the largest training target; on the OOD split, where nearly every query lies above that value, they therefore show what a range-limited model class can achieve.

\paragraph{Closed-form and symbolic baselines.}
To test whether the answers can be recovered from the numbers alone, we add baselines with an arithmetic inductive bias. They are fit on the primary slots only and evaluated at cap 128 on the same subsamples and query rows as the tabular methods (Appendix~\ref{app:rerun}). The closed-form models are ordinary least squares, with and without $\log(1+|x|)$ features; a log-linear model for multiplicative laws; polynomial regression of degree 2, and of degree 3 with a ridge penalty; and degree-2 polynomial regression over the slots and their reciprocals. Symbolic regression uses PySR \citep{cranmer2023pysr} with $+,-,\times,\div$, floor and absolute value and a 120-second time limit per family. Two references complete the set: 5-nearest neighbors and the training maximum, the constant that flat extrapolation returns. Because tied targets can fall on both sides of the split, the training maximum is occasionally exact.

\paragraph{In-context learning.}
GPT-OSS-120B \citep{openai2025gptoss} predicts by in-context learning (ICL). It receives key--value serializations of context rows followed by query rows and a request for numeric predictions; the keys are the column names, and the prompt names the task by its identifier, such as \texttt{gsm8k-000007} (Appendix~\ref{app:prompts}). It runs at temperature zero and row caps 32, 64, and 128, with up to ten attempts to obtain a parseable response; an explicit ``unknown'' reply yields predictions of zero.

\paragraph{Aligned comparisons and uncertainty.}
Every method is run on all 100 GSM8K and 14 AIME families. Cross-method comparisons hold the cap and split fixed and use the families on which both methods have a score, and scaling curves hold each method's family set fixed across caps (Appendix~\ref{app:protocol-coverage}). GPT-OSS-120B's OOD split, like the others', holds out the rows with the largest targets, while its random-split queries come from its own shuffle (Appendix~\ref{app:prompts}). For methods $a$ and $b$, we average $A_{i,a}-A_{i,b}$ over their shared families and bootstrap whole families 20,000 times; bracketed ranges are these 95\% intervals. We add Wilcoxon signed-rank tests on the paired family-level accuracies. Because each family contributes only 26 queries at cap 128, we also separate query-sampling noise from variation across families and check that method orderings hold across the seven row caps (Appendix~\ref{app:seeds}).

\section{Results}
Unless noted, results are for the 100 GSM8K families at cap 128.

\subsection{Strong Regression Fit Conceals Low Answer Precision}
TabPFN v2.5 is the strongest of the nine tabular methods, and its predictions follow the targets beyond the training range. Its median $R^2$ is 0.999 on the random split (Table~\ref{tab:full-GSM8K-reg}) and 0.76 on the OOD split, the only positive OOD median among the tabular methods and far above the $\FlatMedianRtwo$ of predicting the training maximum (Table~\ref{tab:baselines}). Its OOD accuracy, however, is 7.8\%, and it answers no OOD query correctly in 70 of the 100 families. Within families, the two scores come apart: TabPFN v2.5 reaches $R^2>0.8$ on \JointHighFit{} families, yet its accuracy is below 20\% on \JointPrecisionGap{} of them. The tree ensembles, which cannot exceed the training maximum, answer 0.0--0.2\% of OOD queries.

\begin{table}[!htbp]
\centering\small
\caption{\textbf{TabPFN v2.5 follows the OOD targets but rarely answers them exactly, while PySR on the primary slots solves most families.} All methods at cap 128: mean rounded accuracy (\%) over families and median $R^2$, on GSM8K unless noted. ``Solved'' counts the GSM8K families in which every OOD query is answered exactly. The AIME column averages the 13 numerical-resolution-stable families. GPT-OSS-120B is averaged over the families with a score (Appendix~\ref{app:protocol-coverage}). Table~\ref{tab:full-GSM8K-reg} gives the tabular methods' random-split $R^2$, and Appendix~\ref{app:protocol-xrfm} describes the corrected xRFM evaluation.}
\label{tab:baselines}
\fitwidth{\begin{tabular}{lrrrrr}
\toprule
Method & Rand. acc. & OOD acc. & OOD med. $R^2$ & Solved & AIME OOD \\
\midrule
\multicolumn{6}{l}{\emph{Tabular foundation and neural models}} \\
TabPFN v2.5 & 39.5 & 7.8 & 0.76 & 2 & 0.0 \\
TabPFN v2 & 25.3 & 3.8 & -1.34 & 0 & 0.3 \\
TabM & 9.8 & 0.3 & -1.18 & 0 & 0.0 \\
RealMLP-TD-S & 4.0 & 0.3 & -4.35 & 0 & 0.0 \\
xRFM & 7.0 & 0.4 & -2.99 & 0 & 0.0 \\
\multicolumn{6}{l}{\emph{Tree ensembles (cannot exceed the training maximum)}} \\
XGBoost & 5.2 & 0.0 & -4.06 & 0 & 0.0 \\
LightGBM & 5.2 & 0.2 & -3.35 & 0 & 0.0 \\
CatBoost & 4.9 & 0.0 & -5.22 & 0 & 0.0 \\
Random Forest & 4.6 & 0.0 & -4.30 & 0 & 4.7 \\
\multicolumn{6}{l}{\emph{Closed-form and symbolic baselines}} \\
OLS & 17.2 & 14.7 & -0.76 & 14 & 7.7 \\
OLS + log features & 17.7 & 14.9 & -0.94 & 14 & 8.0 \\
Log-linear & 25.2 & 21.8 & 0.53 & 16 & 15.7 \\
Polynomial (deg 2) & 42.0 & 38.0 & 0.94 & 34 & 12.7 \\
Polynomial (deg 3, ridge) & 38.8 & 33.2 & 0.99 & 17 & 5.6 \\
Polynomial over slots and reciprocals & 49.3 & 46.4 & 0.98 & 41 & 18.3 \\
PySR & 81.9 & 76.1 & 1.00 & 72 & 15.7 \\
5-NN & 4.7 & 0.1 & -4.03 & 0 & 5.6 \\
Training maximum (flat reference) & 1.0 & 3.9 & -1.67 & 1 & 10.4 \\
\multicolumn{6}{l}{\emph{In-context learning}} \\
GPT-OSS-120B (ICL) & 49.6 & 39.2 & -1.25 & 32 & 18.2 \\
\bottomrule
\end{tabular}
}
\end{table}

Every tabular method is less accurate on the OOD split than on the random split. Pairing each family's two accuracies cancels its overall difficulty, and every tabular method's OOD-minus-random interval lies below zero; TabPFN v2.5 loses 31.7 points [25.9, 37.7] (Table~\ref{tab:shift}). Methods with small losses, such as RealMLP-TD-S and the tree ensembles, start near zero and have little accuracy to lose.

\subsection{More Rows Improve Interpolation, Not Extrapolation}
\label{sec:rows}
\begin{figure}[!htbp]
\centering
\includegraphics[width=\linewidth]{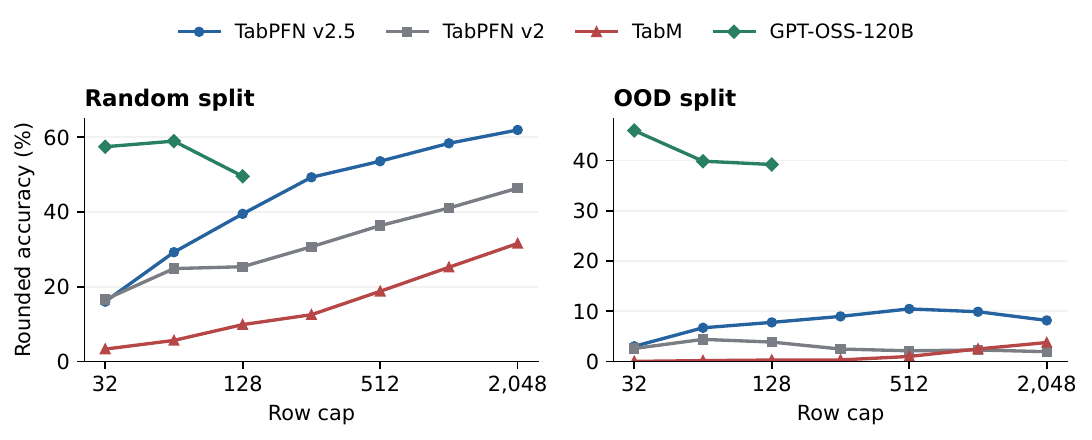}
\caption{\textbf{More rows help the tabular methods on the random split far more than on the OOD split.} GSM8K rounded accuracy; left: random split, right: OOD split. The tabular curves use all 100 families at every cap; GPT-OSS-120B uses one fixed set of families at all three of its caps and on both splits (Appendix~\ref{app:protocol-coverage}). Lines connect observed caps.}
\label{fig:scaling}
\end{figure}
More rows help TabPFN v2.5 interpolate but not extrapolate (Figure~\ref{fig:scaling}). From 32 to 2,048 rows, its random-split accuracy rises from 16.0\% to 62.0\%, while its OOD accuracy rises from 3.0\% to a peak of 10.5\% at 512 rows and falls back to 8.2\%. Paired over the 100 families, moving from 128 to 2,048 rows changes random-split accuracy by $+22.5$ points [18.4, 26.6] and OOD accuracy by $+0.4$ [$-3.4$, 3.9].

The pattern holds across the tabular methods: every other method also gains far more on the random split than on the OOD split as rows grow (Table~\ref{tab:scaling}). A rising OOD $R^2$ need not bring exact answers either. TabM's median OOD $R^2$ becomes positive at 512 rows and reaches 0.455 at 2,048 rows, while its OOD accuracy at that cap is 3.8\%.

\begin{table}[!htbp]
\centering\small
\caption{\textbf{Additional rows raise random-split accuracy, while OOD accuracy stays below 9\%.} GSM8K rounded accuracy of the tabular methods at three of the seven row caps; every cell weights the same 100 families equally.}
\label{tab:scaling}
\begin{tabular}{lrrrrrr}
\toprule
Method & Rand. 32 & 128 & 2,048 & OOD 32 & 128 & 2,048 \\
\midrule
TabPFN v2.5 & 16.0 & 39.5 & 62.0 & 3.0 & 7.8 & 8.2 \\
TabPFN v2 & 16.6 & 25.3 & 46.4 & 2.6 & 3.8 & 1.9 \\
TabM & 3.3 & 9.8 & 31.6 & 0.0 & 0.3 & 3.8 \\
CatBoost & 2.0 & 4.9 & 29.6 & 0.0 & 0.0 & 0.5 \\
XGBoost & 1.6 & 5.2 & 24.0 & 0.1 & 0.0 & 0.4 \\
LightGBM & 2.3 & 5.2 & 23.6 & 0.0 & 0.2 & 0.3 \\
Random Forest & 2.3 & 4.6 & 17.9 & 0.0 & 0.0 & 0.1 \\
RealMLP-TD-S & 2.0 & 4.0 & 19.3 & 0.1 & 0.3 & 1.0 \\
xRFM & 2.7 & 7.0 & 36.8 & 0.0 & 0.4 & 1.8 \\
\bottomrule
\end{tabular}

\end{table}

\subsection{Symbolic Regression Recovers Most Exact Answers}
\label{sec:baselines}
By construction, the inputs determine every answer; the question is whether 102 training rows reveal the rule. For most GSM8K families they do. On the OOD split, \BestBaselineName{}, which sees only the primary slots, reaches \BestBaselineShift\% accuracy and solves \BestBaselineSolved{} of the 100 GSM8K families, whereas TabPFN v2.5 reaches 7.8\% and solves 2 (paired difference $+68.3$ points [59.9, 76.5], Wilcoxon $p=2.9\times10^{-15}$; Table~\ref{tab:baselines}). The input shift that accompanies the output shift costs PySR little: its accuracy averages 76.5\% on OOD queries whose primary slots all lie inside their training ranges and 69.8\% on the others, and no closed-form or symbolic baseline loses more than 9 points on out-of-range queries (Table~\ref{tab:inrange}).

\begin{table}[!htbp]
\centering\small
\caption{\textbf{Tabular foundation models miss computations that a short closed-form fit recovers.} OOD GSM8K accuracy (\%) at cap 128, by computation type. We classify each family by the simplest closed-form model that reproduces its rows: using 5-fold cross-validation on all 2,048 rows of the primary slots, we try linear, multiplicative (log-linear), quadratic, rational (degree-2 polynomial over slots and reciprocals) and cubic models in that order, and assign the first type whose model returns the exact rounded target on at least 99\% of held-out rows. No family needs the cubic model. Families that none of these five models reproduces are labeled \emph{other} (for example, computations with conditionals, integer operations, iteration, search or roots). GPT-OSS-120B is averaged over the families with a score.}
\label{tab:types}
\fitwidth{\begin{tabular}{lrrrrr}
\toprule
OOD accuracy & linear & multiplicative & quadratic & rational & other \\
\midrule
Families & 14 & 11 & 19 & 7 & 49 \\
\midrule
TabPFN v2.5 & 33.5 & 0.3 & 6.5 & 17.6 & 1.2 \\
TabPFN v2 & 15.7 & 0.3 & 2.0 & 13.7 & 0.5 \\
TabM & 1.6 & 0.0 & 0.0 & 0.0 & 0.1 \\
RealMLP-TD-S & 0.5 & 0.0 & 0.2 & 0.0 & 0.3 \\
xRFM & 1.6 & 0.0 & 0.2 & 0.5 & 0.2 \\
XGBoost & 0.0 & 0.0 & 0.0 & 0.0 & 0.1 \\
LightGBM & 1.4 & 0.0 & 0.0 & 0.5 & 0.0 \\
CatBoost & 0.0 & 0.0 & 0.0 & 0.0 & 0.0 \\
Random Forest & 0.0 & 0.0 & 0.0 & 0.0 & 0.0 \\
OLS & 100.0 & 0.0 & 0.0 & 2.2 & 1.1 \\
OLS + log features & 100.0 & 0.0 & 0.8 & 1.6 & 1.3 \\
Log-linear & 28.3 & 100.0 & 2.0 & 8.8 & 11.9 \\
Polynomial (deg 2) & 100.0 & 9.1 & 100.0 & 4.4 & 7.6 \\
Polynomial (deg 3, ridge) & 82.1 & 2.1 & 55.7 & 26.9 & 18.4 \\
Polynomial over slots and reciprocals & 100.0 & 18.2 & 95.1 & 100.0 & 10.9 \\
PySR & 100.0 & 100.0 & 82.0 & 100.0 & 58.2 \\
5-NN & 0.0 & 0.0 & 0.0 & 0.0 & 0.2 \\
Training maximum (flat reference) & 3.8 & 0.7 & 2.0 & 5.5 & 5.2 \\
GPT-OSS-120B (ICL) & 36.0 & 81.8 & 52.5 & 14.3 & 26.7 \\
\bottomrule
\end{tabular}
}
\end{table}

Classifying the families by computation type shows which computations the tabular methods miss (Table~\ref{tab:types}). Every linear, multiplicative, quadratic and rational family is answered exactly on the OOD split by at least one closed-form or symbolic method trained only on the 102 lowest-output rows. TabPFN v2.5, by contrast, averages 33.5\% on the linear families and at most 17.6\% on the other three types, and these four types account for 22 of the 33 families with $R^2>0.8$ and accuracy below 20\% (10 quadratic, 7 multiplicative, 4 linear and 1 rational). Tabular foundation models thus miss computations that a short fit recovers from 102 rows. The open frontier lies elsewhere: 23 of the 49 GSM8K families of other type are solved by no closed-form or symbolic method, and the AIME families remain hard for every method (Section~\ref{sec:aime}).

\subsection{In-Context Learning Answers All or Nothing}
\label{sec:llm}
In-context learning with GPT-OSS-120B answers far more OOD queries than any tabular method. Compared on the same families, it scores 39.2\% on the OOD split against 8.9\% for TabPFN v2.5, a 30.3-point difference [19.1, 41.6] (Wilcoxon $p=3.1\times10^{-5}$), whereas on the random split the two are close ($+5.5$ points [$-8.4$, 19.4], $p=0.34$). The OOD split costs it 10.3 points [0.6, 20.0] (Table~\ref{tab:shift}).

On the OOD split, its accuracy is all-or-nothing per family (Table~\ref{tab:llm-pattern}): it answers every query exactly in 32 families and none in 49, and 25 families are perfect on both splits. In the 49 failed families its errors are large (median $R^2=-11.1$), which produces its negative overall median $R^2$ (Table~\ref{tab:baselines}). For comparison, \BestBaselineName{} solves 68 of the same families.

\subsection{Competition Mathematics Remains Open}
\label{sec:aime}
\begin{figure}[!htbp]
\centering
\includegraphics[width=0.8\linewidth]{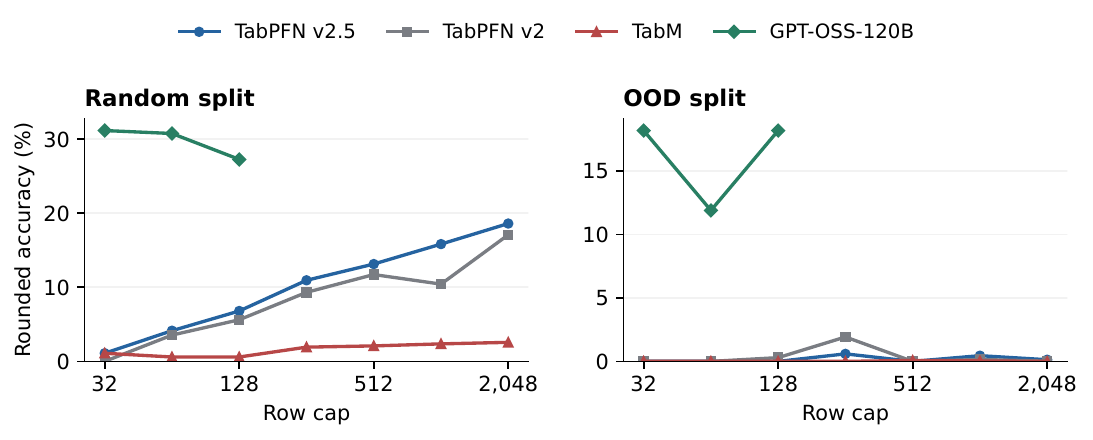}
\caption{\textbf{On AIME, OOD accuracy stays low for every method.} Rounded accuracy with each family weighted equally; left: random split, right: OOD split. The tabular curves use the 13 numerical-resolution-stable families; GPT-OSS-120B uses a fixed set of these families at all three of its caps (Table~\ref{tab:coverage}). Figure~\ref{fig:full-aime} shows every method.}
\label{fig:aime}
\end{figure}
On AIME, no tabular method exceeds 7.7\% OOD accuracy at any row cap (Table~\ref{tab:full-AIME-ood}), and TabPFN v2.5 stays below 1\% (Figure~\ref{fig:aime}); additional rows mainly improve accuracy on the random split. The closed-form and symbolic baselines do not change this picture: at cap 128, the best of them reaches 18.3\% OOD accuracy on these families, and GPT-OSS-120B reaches 18.2\% (Table~\ref{tab:baselines}). Appendix~\ref{app:aime-example} works through one family in detail: PySR's search finds no exact formula there and answers 3.8\% of queries on either split, while only the polynomial over slots and reciprocals answers most OOD queries.

\FloatBarrier

\section{What the Benchmark Reveals}
\paragraph{Four behaviors on the same tables.} Given the same input--output rows, the evaluated methods behave in four distinct ways. \emph{Range-limited} predictors, the tree ensembles, cannot output values above the training maximum and answer almost no OOD query. \emph{Trend-following} predictors track the targets but miss the exact values: TabPFN v2.5 keeps a positive median OOD $R^2$ yet answers no OOD query in most families. In-context learning with GPT-OSS-120B is \emph{all-or-nothing}: in most of its families it answers every OOD query or none. \emph{Exact-recovery} methods fit a formula that matches the computation, as PySR does in most GSM8K families; where its search finds no exact formula, as on the AIME family of Appendix~\ref{app:aime-example}, it answers almost no query. These behaviors reflect inductive bias rather than sample size: sixteen times more rows barely raise OOD accuracy, and the same tables yield exact answers to other methods. $R^2$ separates a small miss from an extreme error, rounded accuracy separates trend-following from exact recovery, and the two scores are complementary.

\paragraph{What the benchmark isolates.} Tabular models are built to interpolate robustly on noisy data, a desirable bias in many real-world settings. \bench{} isolates where that strength becomes a liability: deterministic computations whose exact answers lie beyond the training range, under only a mild shift of the inputs (Section~\ref{sec:evaluation}). Because each family is a separate computation, the benchmark reports not only averages but also which kinds of computation a method recovers (Table~\ref{tab:types}).

\paragraph{Toward predictors that learn the rule.}
These behaviors point to a concrete target: predictors that combine the robustness of tabular foundation models with the exactness of symbolic fits, whether through program induction, arithmetic modules, or hybrid architectures. Because fit alone cannot show that a model has recovered the rule, \bench{} tests such predictors through answer accuracy, error magnitude, and improvement as the row cap grows. The families that no closed-form or symbolic method solves (Section~\ref{sec:baselines}) and the AIME families mark the open frontier.

\section{Related Work}
\paragraph{Tabular prediction.}
Comparative studies relate the performance of tree ensembles and neural predictors to dataset properties, tuning, and computation \citep{grinsztajn2022tree,shwartz2022tabular,mcelfresh2023neural,borisov2022deep}. Neural architectures learn differentiable ensembles of oblivious decision trees \citep{popov2020node}, select or contextualize features \citep{arik2021tabnet,huang2020tabtransformer}, improve feature representations and attention \citep{gorishniy2021revisiting,gorishniy2022embeddings,somepalli2021saint}, attend across datapoints \citep{kossen2021npt}, or form parameter-efficient ensembles \citep{gorishniy2024tabm}; pretraining objectives offer another way to transfer useful structure \citep{rubachev2022pretraining}. Shared infrastructure such as OpenML \citep{vanschoren2014openml} and scikit-learn \citep{pedregosa2011scikit} standardizes datasets, tasks, and estimator interfaces. \bench{} gives these approaches a common deterministic target: every row of a family comes from one executable relationship, so evaluation can measure statistical fit and exact answers separately, both within and beyond the training range.

\paragraph{Learned priors and in-context learning.}
Prior-fitted networks learn a prediction algorithm from synthetic task distributions \citep{muller2022prior,hollmann2023tabpfn,hollmann2025tabpfn,tabpfn25}, and \citet{zhang2025tabpfnone} analyze TabPFN v2 as a statistical estimator. Studies of in-context learning ask which functions and learning procedures a transformer can recover from examples \citep{garg2022context,akyurek2023learning}, and Bayesian and gradient-based accounts analyze this behavior in structured settings \citep{xie2022bayesian,vonoswald2023gradient}. \bench{} poses the same question on computations drawn from human-written problems: observe rows from a fixed computation and return its outputs on new inputs. On these computations we find that TabPFN v2.5 tracks the trend beyond the training range yet rarely returns the exact answer.

\paragraph{Extrapolation and arithmetic.}
Systematic generalization benchmarks such as SCAN, COGS, and CFQ test the recombination of known elements \citep{lake2018generalization,kim2020cogs,keysers2020cfq}, broader analyses organize compositional generalization along several dimensions \citep{hupkes2020compositionality}, and computation-graph analyses connect reasoning difficulty to the composition of intermediate subprocedures \citep{dziri2023faith}. Studies of transformer arithmetic show that number formatting, training examples, and positional representations materially affect generalization \citep{nogueira2021arithmetic,lee2023arithmetic,jelassi2023length,mcleish2024arithmetic}, and work on neural-network extrapolation connects behavior outside the training distribution to architectural and feature-level structure \citep{xu2021extrapolate}. Neural arithmetic logic units (NALU) and neural arithmetic units (NAU) build addition and multiplication into learnable modules \citep{trask2018nalu,madsen2020nau}. \bench{} keeps a family's computation fixed and queries a different region of its output range, an axis of generalization distinct from composition and from the digit-length shifts of sequence-based arithmetic.

\paragraph{Symbolic regression and program synthesis.}
Symbolic regression searches for an explicit expression, as in PySR and AI Feynman \citep{cranmer2023pysr,udrescu2020aifeynman}. AI Feynman 2.0 exploits modular structure \citep{udrescu2020feynman2}, deep symbolic regression trains a policy to search over expressions \citep{petersen2021deepregression}, neural and end-to-end transformer approaches learn to propose equations from observed values \citep{biggio2021symbolic,kamienny2022endtoend}, and symbolic distillation extracts mathematical relationships from learned graph-network representations \citep{cranmer2020discovering}. SRBench evaluates both predictive performance and recovery of known equations \citep{lacava2021srbench}, and program synthesis extends the representation to executable procedures \citep{austin2021program,chen2021codex}. \bench{} scores the numerical answers on held-out rows, so methods that represent the same computation differently are compared on one criterion; PySR serves as our symbolic baseline.

\paragraph{Mathematical reasoning in language.}
Mathematical benchmarks pose problems in natural language \citep{hendrycks2021math,amini2019mathqa,patel2021svamp,dua2019drop}: MathQA associates word problems with operation-based representations, SVAMP probes robustness to variations in elementary problems, and DROP requires discrete reasoning over text. Minerva, Llemma, and DeepSeekMath develop mathematical capabilities through specialized training data and objectives \citep{lewkowycz2022minerva,azerbayev2024llemma,shao2024deepseekmath}, while PAL and Program of Thoughts connect language reasoning to executable solutions \citep{gao2023pal,chen2023program}. GSM-Symbolic shows that language-model accuracy drops when only the numbers in templated GSM8K problems change \citep{mirzadeh2025gsmsymbolic}, and GSM-Ranges probes numerical-scale robustness \citep{shrestha2025gsmranges}. In these benchmarks the model reads the problem, whereas a \bench{} learner sees only input--output rows. In concurrent work, VeRA \citep{cheng2026vera} synthesizes verified reasoning problems from generator--verifier pairs to train and evaluate language models; \bench{} instead compiles generator--verifier programs into tabular regression tasks. An unrelated benchmark also named \bench{} \citep{tian2026tabularmath} evaluates language models reasoning over tables built from word problems.

\section{Conclusion}
We introduced \bench{}, a benchmark for computational extrapolation in tabular learning built from 114 mathematical problems compiled into verified programs. Our central finding is that regression fit and exact answers come apart. With 128 rows and nearly every answer beyond the training range, TabPFN v2.5 is the only tabular method whose median $R^2$ stays positive, yet its OOD accuracy is 7.8\%, and sixteen times more rows do not close the gap. Every answer is determined by its inputs by construction, and most are recoverable from the same rows: symbolic regression on the numbers alone solves 72 of 100 GSM8K families, and in-context learning with GPT-OSS-120B answers 39.2\% of OOD queries. The failure of tabular foundation models is one of inductive bias, not of information. Competition-level AIME computations remain open for every method.

In applications that need exact outputs, such as pricing, simulation, and business rules, a high $R^2$ is no evidence that a model has learned the rule. \bench{} offers a direct test for predictors that combine statistical robustness with program-like exactness: exact answers beyond the training range, scored family by family.
\label{sec:main-end}

\bibliographystyle{\bibstylename}
\bibliography{references}
\clearpage
\appendix\raggedbottom
\section{Specification Construction}
\label{app:construction-details}
This appendix specifies what the compiler produces for each seed, the checks that every candidate specification must pass, and the features derived from each row. Appendix~\ref{app:worked} follows an illustrative family through these steps, Appendix~\ref{app:records} describes seed selection and table assembly, and Appendix~\ref{app:protocol} gives the evaluation protocol.

\subsection{The Compiler's Contract}
The compiler is GPT-5 with high reasoning effort \citep{openai2025gpt5}. The synthesis prompt asks it to generalize a seed while preserving its calculation. The compiler returns a single structured specification with five linked objects: text templates, typed slots, a generator, a verifier, and the seed's base assignment. A slot can represent a numerical quantity, an entity, a unit, or a categorical choice. The same slot names must appear in the templates, schema, generator output, and verifier input. This agreement matters because a program can execute correctly while a rendered question refers to a different quantity.

The generator, verifier, and templates have fixed interfaces. The generator accepts a supplied random-number source and returns an assignment. The verifier accepts that assignment and returns a validity flag and a numeric answer. The templates use the assignment to express the question without exposing its answer. For tabular evaluation, the input columns are derived from the assignment and its generic features, and the target is obtained by executing the verifier. Learners observe only these rows; the reference program remains part of the benchmark construction.

\subsection{Construction Checks}
Table~\ref{tab:construction-checks} lists the checks that a candidate specification must pass and the failure each one detects. When a check fails, the feedback identifies the concrete error, such as a missing slot, a failed constraint, or a mismatch with the seed answer, and the next proposal can revise the affected generator, verifier, or template.

\begin{table}[H]
\centering\small
\caption{\textbf{Every candidate specification must pass five checks.} Each check and the failure it detects. A candidate is revised when a check fails, within the 20-attempt synthesis budget.}
\label{tab:construction-checks}
\begin{tabularx}{\linewidth}{lY}
\toprule
Check & Failure detected\\
\midrule
Schema agreement & A template requests a slot that the generator omits, or code uses an undeclared input.\\
Seed anchoring & The verifier fails to recover the published answer at the original assignment.\\
Valid generation & Sampled counts, denominators, or other quantities violate the task's constraints.\\
Deterministic execution & The same assignment receives different labels, or labeling fails to terminate within its time limit.\\
Table integrity & Duplicate rows (compared with their text metadata), nonnumeric targets, or malformed feature records enter the learning task.\\
\bottomrule
\end{tabularx}
\end{table}

\subsection{Feature Representation}
Each row combines the primary numerical slots with derived features: log-magnitude $\log(1+|x|)$, sign, parity, remainders for small moduli, fractional part and number of decimals for real-valued slots, categorical encodings where applicable, and two divisibility summaries across the integer slots (their greatest common divisor and the number of slot pairs in which one divides the other). These generic transformations expose common numerical structure, but none of them reproduces a family's answer: across all 114 tables, no derived feature equals the rounded target on more than 29\% of rows. An analysis of a particular family can inspect its column definitions alongside the verifier to identify which dependencies the learner must recover.

\section{A Worked Generator\textendash{}Verifier Family}
\label{app:worked}
This appendix applies the construction of Appendix~\ref{app:construction-details} to an illustrative flower-counting family: it follows the family from its seed problem to labeled rows, shows how the random and OOD splits divide those rows, and contrasts it with an example whose answers are continuous. This family is illustrative and not one of the 114 benchmark families; Appendix~\ref{app:aime-example} works through a benchmark family.

\subsection{From Seed to Labeled Rows}
A garden has $a$ yellow flowers, $b$ percent more purple flowers than yellow flowers, and green flowers numbering $c$ percent of the combined yellow and purple count. The seed uses $(a,b,c)=(10,80,25)$ and has answer 35. Parameterization turns the three quantities into slots while retaining the relations between them:
\begin{equation}
 p=a(1+b/100),\qquad g=(a+p)c/100,\qquad y=a+p+g.
\end{equation}

The generator proposes $a\in\{1,\ldots,100\}$ and $b,c\in\{0,\ldots,100\}$ and accepts an assignment only if both $p$ and $g$ are integers. These domain constraints are part of the task: they preserve the meaning of counting flowers, rather than producing fractional objects and rounding away the inconsistency. The verifier evaluates each accepted assignment, and the resulting row contains the inputs and their computed target. Table~\ref{tab:worked-family} shows five such rows.

\begin{table}[H]
\centering
\caption{\textbf{A seed becomes a family of labeled rows.} Assignments of the flower-counting example before the generic feature transforms. All rows follow the same generator constraints and answer program; the final row is the original seed.}
\label{tab:worked-family}
\begin{tabular}{rrrrrr}
\toprule
Yellow $a$ & Extra purple (\%) & Green (\%) & Purple $p$ & Green $g$ & Target $y$\\
\midrule
20 & 50 & 40 & 30 & 20 & 70\\
15 & 100 & 20 & 30 & 9 & 54\\
40 & 50 & 12 & 60 & 12 & 112\\
8 & 25 & 50 & 10 & 9 & 27\\
10 & 80 & 25 & 18 & 7 & 35\\
\bottomrule
\end{tabular}
\end{table}

\subsection{How the Splits Divide the Rows}
The rows in Table~\ref{tab:worked-family} change the counts while preserving the same three relations. A random split samples training and query rows throughout this distribution. The OOD split instead sorts rows by the answer, so it trains on smaller totals and queries larger ones. The executable family keeps the relations fixed across both splits.

Because different combinations of inputs can produce the same total, a larger answer need not require every input to be larger. The row with target 112 has more yellow flowers than the row with target 70, but a smaller green percentage, 12\% rather than 40\%; its larger answer comes from the combined calculation, not from increasing every coordinate. Individual input coordinates can therefore overlap between training and query rows, while the query rows occupy the upper end of the output distribution. To answer them, a predictor must account for how the quantities interact.

\subsection{A Continuous-Quantity Example}
Not every family has integer-valued answers. A distance--rate--time example shows how the generator samples coherently when quantities depend on one another. Let $v$ be base speed, $h$ a headwind reduction, and $t$ elapsed time. The generator samples $v\in[6,10]$, $h\in[1,5]$, and $t\in[1,10]$, then sets distance $d=(v-h)t$. A row exposes $(d,h,t)$, and its label
\[
 V(d,h,t)=d/t+h
\]
recovers the base speed, which can be nonintegral. The intervals ensure positive effective speed and a nonzero denominator. The original example $(d,h,t)=(15,3,5)$ has answer 6.

Constructing the distance from $v$, $h$, and $t$ preserves the intended relation directly; sampling it independently of the other quantities would require additional validity checks. The two examples place different requirements on valid generation (integrality constraints for the flowers, coherent continuous arithmetic here), and both support deterministic labels. The continuous case motivates defining the scoring resolution explicitly rather than assuming that every program returns a count (Section~\ref{sec:evaluation}).

\section{Seed Selection and Table Construction}
\label{app:records}
This appendix records how the seeds were chosen, when a seed is accepted, how each table's rows are assembled, and how many rows repeat within a table.

\paragraph{Seeds.} The 100 GSM8K seeds were drawn uniformly at random from the GSM8K test split, which has 1,319 problems. Family \texttt{gsm8k-}$n$ is problem $n$ of that split, counting from zero; the indices range from 7 to 1293. The 14 AIME seeds are the problems of AIME 2024 (30 in total) whose families yield at least 2,048 distinct rows during construction; the table-building configuration lists them. By the classification of Table~\ref{tab:types}, the AIME families comprise 11 computations of other type and one each linear, multiplicative and rational.

\paragraph{Acceptance.} A seed is accepted once its specification passes the checks of Table~\ref{tab:construction-checks}. Seeds without a passing generator--verifier pair after 20 synthesis attempts are discarded.

\paragraph{Row assembly.} Table rows come first from the variants sampled during synthesis and then from fresh generator draws labeled by the verifier; draws that the verifier rejects are skipped. During synthesis, the sampler falls back to the base assignment if 100 draws fail, and such a draw reproduces the seed row.

\paragraph{Duplicates.} Rows are deduplicated while they still carry text-template metadata (question length and perturbation scores), which the benchmark tables omit. As a result, 27 of the 233,472 benchmark rows repeat another row of the same table, in 11 of the 114 tables.

\section{A Worked AIME Family}
\label{app:aime-example}
This appendix follows one benchmark family, 2024-I-1, the first in the AIME list, from its seed problem to every method's result at a 128-row cap.

\paragraph{The seed.} The seed (AIME 2024 I, Problem 1), paraphrased: a walker covers a fixed distance and then spends the same number of minutes in a coffee shop; at speed $s$ the outing takes $T_1$ hours, at speed $s+\Delta$ it takes $T_2$ hours, and the question asks how many minutes it takes at speed $s+\frac12$. In the seed, $d=9$\,km, $\Delta=2$\,km/h, $T_1=4$\,h and $T_2=2.4$\,h, and the answer is 204. The family has four primary slots ($d$, $\Delta$, $T_1$, $T_2$) and three categorical slots (the walker's name, the activity and the venue).

\paragraph{The computation.} To compute the answer, one solves $d/s-d/(s+\Delta)=T_1-T_2$, a quadratic in $s$, and then evaluates
\[
 y=60\Bigl(\frac{d}{s+1/2}+T_1-\frac{d}{s}\Bigr),\qquad s=\tfrac12\Bigl(-\Delta+\sqrt{\Delta^2+4d\Delta/(T_1-T_2)}\Bigr).
\]
All 2,048 rows of the family satisfy this formula to within $10^{-12}$.

\paragraph{The rows.} The family's rows show the generator's range: $d$ takes 23 integer values from 6 to 56, $\Delta\in\{1,2,3\}$, the implied speed $s$ is one of $1.5, 2.5,\dots,6.5$, and the implied coffee-shop time is one of $6, 12,\dots,42$ minutes, so every target is an integer between 186 and 522. Table~\ref{tab:aime-rows} shows the first two rows of the family and the seed, which also appears among the rows.
\begin{table}[H]
\centering\small
\caption{\textbf{The seed is itself a row of family 2024-I-1.} The first two rows of the family, followed by the seed (primary slots and target; times rounded to three decimals).}
\label{tab:aime-rows}
\begin{tabular}{rrrrr}
\toprule
Distance $d$ (km) & Offset $\Delta$ (km/h) & $T_1$ (h) & $T_2$ (h) & Target $y$ (min)\\
\midrule
15 & 3 & 6.700 & 3.427 & 342\\
42 & 1 & 6.862 & 6.000 & 384\\
9 & 2 & 4.000 & 2.400 & 204\\
\bottomrule
\end{tabular}
\end{table}

\paragraph{Results at a 128-row cap.} Table~\ref{tab:aime-example} gives every method's result. The OOD queries have targets from 462 to 522, and every one lies at or above the largest training target. No closed-form or symbolic baseline includes the square root in the formula for $s$ in its function class, yet on this discrete input domain the degree-2 polynomial over slots and reciprocals answers 76.9\% of OOD queries exactly (96.2\% on the random split). It also reproduces 99\% of held-out rows in the cross-validation that types the family as rational (Table~\ref{tab:types}). TabPFN v2.5 answers 53.8\% of random-split queries and none on the OOD split, PySR's selected expressions answer 3.8\% on both splits, and GPT-OSS-120B answers 11.5\% of random-split queries and none on the OOD split.
\begin{table}[H]
\centering\small
\caption{\textbf{On family 2024-I-1, only the polynomial over slots and reciprocals answers most OOD queries.} Rounded accuracy (\%) on the 26 queries of each split and OOD $R^2$, at a 128-row cap.}
\label{tab:aime-example}
\begin{tabular}{lrrr}
\toprule
Method & Random acc. & OOD acc. & OOD $R^2$ \\
\midrule
TabPFN v2.5 & 53.8 & 0.0 & -0.29 \\
TabPFN v2 & 46.2 & 3.8 & 0.74 \\
TabM & 3.8 & 0.0 & -0.05 \\
RealMLP-TD-S & 3.8 & 0.0 & -13.52 \\
xRFM & 15.4 & 0.0 & -17.96 \\
XGBoost & 0.0 & 0.0 & -17.51 \\
LightGBM & 0.0 & 0.0 & -18.59 \\
CatBoost & 0.0 & 0.0 & -27.89 \\
Random Forest & 0.0 & 0.0 & -18.10 \\
OLS & 11.5 & 0.0 & 0.54 \\
OLS + log features & 19.2 & 3.8 & 0.49 \\
Log-linear & 7.7 & 3.8 & 0.81 \\
Polynomial (deg 2) & 11.5 & 3.8 & 0.88 \\
Polynomial (deg 3, ridge) & 84.6 & 11.5 & 0.98 \\
Polynomial over slots and reciprocals & 96.2 & 76.9 & 1.00 \\
PySR & 3.8 & 3.8 & 0.90 \\
5-NN & 0.0 & 0.0 & -31.16 \\
Training maximum & 7.7 & 3.8 & -9.43 \\
GPT-OSS-120B (ICL) & 11.5 & 0.0 & -49.36 \\
\bottomrule
\end{tabular}

\end{table}

\section{Evaluation Protocol}
\label{app:protocol}
This appendix specifies how the methods are evaluated: the splits and model configurations, the protocol of the closed-form and symbolic baselines, a correction to xRFM's recorded accuracy, the prompt and parsing procedure for GPT-OSS-120B, and the family coverage behind each comparison.

\subsection{Splits and Model Configurations}
\label{app:protocol-splits}
Each family has 2,048 rows, deduplicated at construction while they still carry text-template metadata (Appendix~\ref{app:records}). The evaluated caps are 32, 64, 128, 256, 512, 1,024, and 2,048 total rows. Each table is subsampled to the cap before the 80/20 split, with sampling seed 2025. The random split assesses prediction within the sampled distribution. The OOD split sorts by the target before splitting, which puts the largest outputs in the query set; ties at the boundary remain possible. Both splits retain the same underlying program, but they expose different regions of its output distribution.

Standardization uses training-split statistics. All 114 tables have complete features, so no values are filled. TabPFN v2.5 uses the \texttt{v2.5\_default} API configuration with eight estimators, and the other tabular methods use their fixed evaluation configurations. GPT-OSS-120B receives serialized context and query rows at caps 32, 64, and 128, with temperature zero; Appendix~\ref{app:prompts} gives its prompt and parsing procedure.

\subsection{Closed-Form and Symbolic Baselines}
\label{app:rerun}
The closed-form and symbolic baselines use only the primary numeric slots, that is, the integer- and real-valued slots, without categorical codes or derived features, so they show what those slots support. They take their rows, splits and scoring from the benchmark's evaluation code and therefore see exactly the same 128-row subsamples and query rows as the tabular methods. The recorded results of the other methods are used unchanged, apart from xRFM's accuracy (Appendix~\ref{app:protocol-xrfm}).

\paragraph{Rows and splits.} Each table is subsampled with \texttt{DataFrame.sample(128, random\_state=2025)}, and constant columns are dropped. The random split is the evaluation code's 80/20 \texttt{train\_test\_split} with seed 2025, and the OOD split holds out the 26 rows with the largest targets. Tied targets can straddle this boundary, which is why the training-maximum reference is occasionally exact. Because the split sorts the targets with numpy's default, unstable sort, the order of tied targets depends on the numpy version. The baselines were run with numpy 1.26.4, with which the evaluation driver reproduces the recorded Random Forest metrics exactly on both splits for the three families checked; Appendix~\ref{app:protocol-xrfm} reports its agreement with the xRFM records across all 114 families. Rounded accuracy is computed as in the evaluation code, and non-finite predictions count as wrong.

\paragraph{Models.} In the order of Table~\ref{tab:baselines}, the baselines are:
\begin{itemize}
\item \emph{OLS}: ordinary least squares;
\item \emph{OLS + log features}: least squares on the slots and $\log(1+|x|)$;
\item \emph{Log-linear}: least squares on $\log x$ predicting $\log y$, used when all training values are positive and otherwise replaced by ordinary least squares;
\item \emph{Polynomial (deg 2)}: degree-2 polynomial features with least squares;
\item \emph{Polynomial (deg 3, ridge)}: degree-3 polynomial features with ridge regression, choosing the penalty by leave-one-out;
\item \emph{Polynomial over slots and reciprocals}: degree-2 polynomial features over the slots and the reciprocals of slots that are never zero in training;
\item \emph{PySR}: symbolic regression over $+,-,\times,\div$, floor and absolute value, with at most 30 nodes, 60 iterations, a 120-second limit per family and deterministic serial search, selecting the expression with the lowest training loss;
\item \emph{5-NN}: 5-nearest neighbors on standardized slots;
\item \emph{Training maximum}: a constant equal to the largest training target.
\end{itemize}
OLS, the two plain polynomial models and 5-NN are fit on inputs and targets standardized with training statistics, as the tabular methods are. The log-feature, log-linear and reciprocal models, PySR and the training-maximum reference use unstandardized values.

\subsection{Correction to xRFM's Accuracy}
\label{app:protocol-xrfm}
\paragraph{The error.} The evaluation driver receives xRFM's predictions as an $n\times1$ column and compares it with the $n$ query targets by broadcasting, so the recorded xRFM accuracy averages over all $n^2$ prediction--target pairs. The records show this: 849 of xRFM's 860 nonzero accuracies are multiples of $1/n^2$ but not of $1/n$, whereas every nonzero accuracy of every other method is a multiple of $1/n$. $R^2$ and MSE, computed with scikit-learn, are unaffected.

\paragraph{The corrected rerun.} We reran xRFM (version 0.4.3, as the evaluation code pins) through the same evaluation driver in the environment of Appendix~\ref{app:rerun}, changing only the shape of its predictions. Every xRFM accuracy in this paper comes from the corrected rerun, while its $R^2$, like every other method's result, is the recorded result.

\paragraph{Agreement with the records.} Run without that change, the driver reproduces the recorded xRFM accuracy exactly on 1,311 of its 1,596 records: 219 of 228 at cap 128, falling to 95 of 228 at cap 2,048. The rerun's $R^2$ agrees with the record to within $10^{-3}$ on 1,388 records (median difference $8\times10^{-5}$). The remaining differences come from floating-point differences between software builds, which move some predictions across a rounding boundary.

\paragraph{Effect on the results.} The correction changes the GSM8K values in Table~\ref{tab:xrfm-correction}; no conclusion changes.
\begin{table}[H]
\centering\small
\caption{\textbf{Correcting xRFM's accuracy changes no conclusion.} GSM8K values computed from the recorded xRFM results and from the corrected rerun: xRFM's rounded accuracy, its OOD-minus-random contrast at cap 128 (Table~\ref{tab:shift}) with its 95\% interval, and the median Kendall's $\tau$ between method rankings across pairs of row caps (Appendix~\ref{app:seeds}).}
\label{tab:xrfm-correction}
\begin{tabular}{llrr}
\toprule
Quantity & Row cap & Recorded & Corrected\\
\midrule
Random-split accuracy (\%) & 128 & 1.7 & 7.0\\
 & 2,048 & 1.7 & 36.8\\
OOD accuracy (\%) & 128 & 0.5 & 0.4\\
 & 2,048 & 2.0 & 1.8\\
OOD minus random (pp) & 128 & $-1.2$ [$-2.0$, $-0.5$] & $-6.6$ [$-8.7$, $-4.7$]\\
Median Kendall's $\tau$, random split & -- & 0.82 & 0.76\\
Median Kendall's $\tau$, OOD split & -- & 0.70 & 0.74\\
\bottomrule
\end{tabular}
\end{table}

\subsection{Language-Model Prompt and Parsing}
\label{app:prompts}
GPT-OSS-120B received the key--value prompt below, unchanged from the evaluation code, together with the system message ``You convert tabular regression prompts into numeric predictions.'' In the template, \texttt{\{name\}} is the task identifier (for example \texttt{gsm8k-000007}), and \texttt{\{cols\}} lists every column that remains after constant columns are dropped, including the derived features. Each context or query line lists these columns as \texttt{column=value} pairs; long lines are wrapped here.
\begin{verbatim}
You are an expert regression model. The dataset '{name}' has
numeric features {cols} and target y.
First, learn from the context rows (each line labelled CONTEXT).
Then, predict y for each QUERY row.
Return ONLY a JSON list of floats corresponding to the QUERY rows
in order.

Context rows:
CONTEXT: slot_a=..., slot_b=..., ..., y=...
...

Query rows:
QUERY 0: slot_a=..., slot_b=..., ...
...
\end{verbatim}

\paragraph{Parsing.} The harness reads as many numbers as there are queries from a parseable response, taking the first ones that match \texttt{-?\textbackslash d+(?:\textbackslash .\textbackslash d+)?}. A response with insufficient values is retried, up to ten attempts; the harness returns zeros only for a response containing the placeholder \texttt{<unknown>}.

\paragraph{Splits.} The harness draws its own splits. Its random split shuffles the subsample with \texttt{DataFrame.sample(frac=1, random\_state=2025)} and takes the last 20\% as queries, so its random-split queries differ from those of the tabular methods. Its OOD split, like theirs, holds out the rows with the largest targets.

\subsection{Family Coverage and Comparison Sets}
\label{app:protocol-coverage}
The unit of aggregation is a task family. Every method is run on all 100 GSM8K and 14 AIME families. Each of the nine tabular methods has a scored record for every family, split, and row cap. GPT-OSS-120B is run at caps 32, 64, and 128; the evaluation driver keeps a run that ends in an error as a failure without a score, so its scored GSM8K records number 100, 99, and 87 per split at those caps, and its AIME records 14, 14, and 12, of which 11 are numerical-resolution-stable at cap 128. The 13 GSM8K families without a cap-128 score are wide tables (median 52 feature columns, against 29 for the other 87), whose prompts at that cap exceed the model's context window. Counting these 13 families as wrong gives GPT-OSS-120B an OOD accuracy of 34.1\% over all 100 GSM8K families, still far above TabPFN v2.5's 7.8\%.

Comparisons with GPT-OSS-120B use fixed family sets. The direct model comparison at cap 128 uses the 87 GSM8K families with a GPT-OSS-120B score. The row-cap curves restrict GPT-OSS-120B to families present at every plotted cap and in both split conditions; Table~\ref{tab:coverage} lists these family sets.
\begin{table}[H]
\centering
\caption{\textbf{The GPT-OSS-120B scaling curves use fixed family sets.} Families with GPT-OSS-120B records at every plotted row cap and on both splits.}
\label{tab:coverage}
\begin{tabular}{lrl}
\toprule
Source & Shared families & Row caps \\
\midrule
GSM8K & 87 & 32, 64, 128 \\
AIME & 11 & 32, 64, 128 \\
\bottomrule
\end{tabular}

\end{table}

\section{Uncertainty}
\label{app:seeds}
This appendix describes how the paper quantifies uncertainty. Every interval comes from a bootstrap over task families, query sampling within a family is a small part of that uncertainty, and the ordering of the methods holds across the seven row caps.

\subsection{Paired Intervals and Tests}
For a paired contrast between conditions $a$ and $b$ (two methods, two splits or two row caps), let $d_i=A_{i,a}-A_{i,b}$ on family $i$. The point estimate is $\bar d=n^{-1}\sum_i d_i$. Each bootstrap replicate samples $n$ families with replacement and averages their differences, and the 2.5th and 97.5th percentiles of 20,000 replicates form the 95\% interval; a fixed random seed makes the resampling reproducible. Because whole families are resampled, the interval measures variation across families for the evaluated tables and model runs, preserving both the generated rows within a family and the pairing of model outcomes.
\begin{table}[!htbp]
\centering\small
\caption{\textbf{Every method loses accuracy on the OOD split.} OOD-minus-random rounded accuracy at cap 128, paired by family within each method, with 95\% intervals from a bootstrap over families. GPT-OSS-120B uses its 87 families; each tabular method uses all 100. Differences are averaged before rounding.}
\label{tab:shift}
\begin{tabular}{lrrr}
\toprule
Method & Families & Difference (pp) & 95\% interval \\
\midrule
TabPFN v2.5 & 100 & -31.7 & [-37.7, -25.9] \\
TabPFN v2 & 100 & -21.5 & [-26.7, -16.5] \\
TabM & 100 & -9.6 & [-12.6, -6.8] \\
CatBoost & 100 & -4.9 & [-7.0, -3.1] \\
XGBoost & 100 & -5.1 & [-7.1, -3.4] \\
LightGBM & 100 & -4.9 & [-7.0, -3.2] \\
Random Forest & 100 & -4.6 & [-6.4, -3.0] \\
RealMLP-TD-S & 100 & -3.8 & [-5.8, -2.2] \\
xRFM & 100 & -6.6 & [-8.7, -4.7] \\
GPT-OSS-120B & 87 & -10.3 & [-20.0, -0.6] \\
\bottomrule
\end{tabular}

\end{table}

The bracketed ranges throughout the paper are these intervals (Section~\ref{sec:evaluation}). Table~\ref{tab:shift} applies the estimator to the OOD-minus-random contrast at cap 128, and the paired changes from 128 to 2,048 rows in Section~\ref{sec:rows} use the same estimator. The reported $p$-values come from Wilcoxon signed-rank tests on the same paired family-level accuracies.

\subsection{Query-Sampling Noise}
Each family contributes 26 queries at cap 128, so its accuracy carries query-sampling noise. Table~\ref{tab:noise} separates that noise from variation across families. Because it resamples whole families, the bootstrap over families already includes both sources.
\begin{table}[!htbp]
\centering\small
\caption{\textbf{Query sampling is a small part of the uncertainty.} Standard error (percentage points) of the mean accuracy over families at a 128-row cap on GSM8K, where $p_i$ is the accuracy on family $i$ and $N$ the number of families. Query sampling: $\frac{1}{N}\sqrt{\sum_i p_i(1-p_i)/26}$; across families: the standard deviation of family accuracies divided by $\sqrt{N}$.}
\label{tab:noise}
\begin{tabular}{llrrrr}
\toprule
Method & Split & Families & Mean & Query sampling & Across families \\
\midrule
TabPFN v2.5 & Random & 100 & 39.5 & 0.63 & 3.70 \\
TabPFN v2.5 & OOD & 100 & 7.8 & 0.35 & 2.02 \\
GPT-OSS-120B & Random & 87 & 49.6 & 0.36 & 5.07 \\
GPT-OSS-120B & OOD & 87 & 39.2 & 0.11 & 5.23 \\
\bottomrule
\end{tabular}
\end{table}

\subsection{Stability of Method Rankings}
The seven row caps give a further check on how the methods are ordered. TabPFN v2.5 has the highest OOD accuracy among the tabular methods at all seven caps and the highest random-split accuracy at six of seven (Tables~\ref{tab:full-GSM8K-random} and~\ref{tab:full-GSM8K-ood}). Across pairs of caps, the median Kendall's $\tau$ between method rankings is 0.76 on the random split and 0.74 on the OOD split. The caps are nested subsamples of one random permutation, so this is a check of stability across caps rather than a replication with independent seeds.

\section{In-Range and Out-of-Range Queries}
\label{app:inrange}
Sorting by output places nearly every OOD target above the training range, and it also moves the inputs that raise those targets. This appendix measures how far the inputs move and what out-of-range inputs cost the closed-form and symbolic baselines.

\paragraph{What the OOD split holds fixed.} For a deterministic target $y=V(x)$, conditioning on the upper tail of $y$ selects inputs through $V$. Some families reach that tail through one large input; others require a particular combination, a discrete branch, or several interacting constraints. The reference computation therefore stays fixed while the joint distribution of inputs and outputs changes: the OOD split asks a learner to learn from lower-output examples and predict the computation's upper-output region.

An OOD query is \emph{in range} if each of its primary slots lies between that slot's minimum and maximum over the training rows of the same subsample; derived features are not considered. The in-range share is a property of the split and is the same for every method.

At cap 128, 96.2\% of OOD GSM8K queries have a target above every training target, while the in-range share averages 82.3\% over the 100 GSM8K families, 33 of which have every OOD query in range. Over the 13 numerical-resolution-stable AIME families the share averages 74.0\%, and 6 of them have every OOD query in range. The OOD split therefore tests extrapolation in the output under a mild shift of the inputs.

Table~\ref{tab:inrange} splits the accuracy of the closed-form and symbolic baselines by this criterion. PySR's accuracy averages 76.5\% on in-range and 69.8\% on out-of-range queries, and no closed-form or symbolic baseline loses more than 9 points on out-of-range queries.
\begin{table}[H]
\centering\small
\caption{\textbf{Out-of-range inputs cost the closed-form and symbolic baselines little.} OOD GSM8K accuracy (\%) at a 128-row cap, split by whether every primary slot of a query lies within its training range, as it does for 82.3\% of queries. Each column averages the families that have such queries (100 and 67). Rows are the closed-form and symbolic baselines.}
\label{tab:inrange}
\begin{tabular}{lrr}
\toprule
Method & In range & Out of range \\
\midrule
OLS & 14.6 & 15.7 \\
OLS + log features & 14.9 & 15.8 \\
Log-linear & 23.0 & 21.5 \\
Polynomial (deg 2) & 38.1 & 31.5 \\
Polynomial (deg 3, ridge) & 34.7 & 26.0 \\
Polynomial over slots and reciprocals & 46.4 & 43.0 \\
PySR & 76.5 & 69.8 \\
5-NN & 0.1 & 0.0 \\
Training maximum (flat reference) & 5.5 & 2.8 \\
\bottomrule
\end{tabular}

\end{table}

\section{Tolerance Accuracy}
\label{app:tolerance}
Rounded accuracy fixes an absolute answer resolution; tolerance accuracy instead accepts a prediction within a relative error of 0.5\%, 1\% or 5\% of the target. Table~\ref{tab:tolerance} reports both scores for the closed-form and symbolic baselines on OOD GSM8K queries at a 128-row cap. PySR leads under rounding and at every tolerance.
\begin{table}[H]
\centering\small
\caption{\textbf{PySR leads at every tolerance.} OOD GSM8K accuracy (\%) at a 128-row cap under rounding and under relative-error tolerances, for the closed-form and symbolic baselines. A query counts at tolerance $\tau$ if $|\hat y-y|\le\tau\max(1,|y|)$.}
\label{tab:tolerance}
\fitwidth{\begin{tabular}{lrrrr}
\toprule
Method & Rounded & $\le$0.5\% & $\le$1\% & $\le$5\% \\
\midrule
OLS & 14.7 & 14.5 & 15.1 & 21.7 \\
OLS + log features & 14.9 & 14.7 & 15.3 & 22.7 \\
Log-linear & 21.8 & 22.7 & 26.6 & 47.8 \\
Polynomial (deg 2) & 38.0 & 39.9 & 43.1 & 60.5 \\
Polynomial (deg 3, ridge) & 33.2 & 47.0 & 53.6 & 69.8 \\
Polynomial over slots and reciprocals & 46.4 & 48.3 & 52.1 & 66.3 \\
PySR & 76.1 & 80.5 & 81.6 & 86.2 \\
5-NN & 0.1 & 0.1 & 0.1 & 0.2 \\
Training maximum (flat reference) & 3.9 & 4.3 & 4.7 & 15.2 \\
\bottomrule
\end{tabular}
}
\end{table}

\section{GPT-OSS-120B Results by Family}
\label{app:llm-pattern}
This appendix breaks down the cap-128 GSM8K results of GPT-OSS-120B by family. Table~\ref{tab:llm-pattern} cross-tabulates its per-family accuracy on the two splits, and Table~\ref{tab:llm-worst} lists the families where its errors are largest.

Table~\ref{tab:llm-pattern} shows the all-or-nothing pattern of Section~\ref{sec:llm}: on the OOD split, most of its families sit at 0\% or 100\%. GPT-OSS-120B's largest errors do not always fall on families that are hard for TabPFN v2.5: on three of the ten families with GPT-OSS-120B's lowest OOD $R^2$, TabPFN v2.5 answers at least 42\% of queries exactly.
\begin{table}[H]
\centering\small
\caption{\textbf{On the OOD split, GPT-OSS-120B is all-or-nothing per family.} Number of families by accuracy at a 128-row cap, with the random split in rows and the OOD split in columns, on its 87 GSM8K families.}
\label{tab:llm-pattern}
\begin{tabular}{lrrr}
\toprule
Random split $\backslash$ OOD split & 0\% & In between & 100\% \\
\midrule
0\% & 21 & 2 & 1 \\
In between & 20 & 1 & 6 \\
100\% & 8 & 3 & 25 \\
\bottomrule
\end{tabular}
\end{table}

\begin{table}[H]
\centering\small
\caption{\textbf{Some of GPT-OSS-120B's largest errors fall on families that TabPFN v2.5 answers well.} The ten GSM8K families with GPT-OSS-120B's lowest OOD $R^2$ at a 128-row cap, with TabPFN v2.5's OOD accuracy on the same family.}
\label{tab:llm-worst}
\begin{tabular}{lrr}
\toprule
Family & GPT-OSS-120B $R^2$ & TabPFN v2.5 accuracy (\%) \\
\midrule
gsm8k-000348 & $-229.6$ & 0.0 \\
gsm8k-000734 & $-229.0$ & 88.5 \\
gsm8k-000604 & $-163.4$ & 0.0 \\
gsm8k-000905 & $-159.9$ & 100.0 \\
gsm8k-001031 & $-113.3$ & 3.8 \\
gsm8k-000361 & $-80.0$ & 0.0 \\
gsm8k-000967 & $-72.8$ & 0.0 \\
gsm8k-001236 & $-57.4$ & 11.5 \\
gsm8k-000819 & $-51.9$ & 0.0 \\
gsm8k-001128 & $-49.0$ & 42.3 \\
\bottomrule
\end{tabular}
\end{table}

\section{Complete Results}
\label{app:complete-metrics}
This appendix collects the complete results across methods and row caps: scaling profiles over all available families, and tables of rounded accuracy at all seven row caps and of regression errors at caps 128 to 2,048.

\subsection{Scaling Profiles over All Families}
\begin{figure}[tp]
\centering
\includegraphics[width=\linewidth]{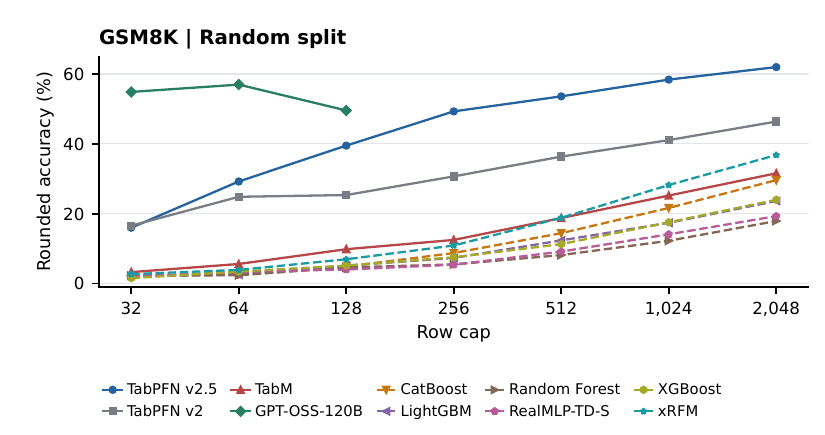}
\includegraphics[width=\linewidth]{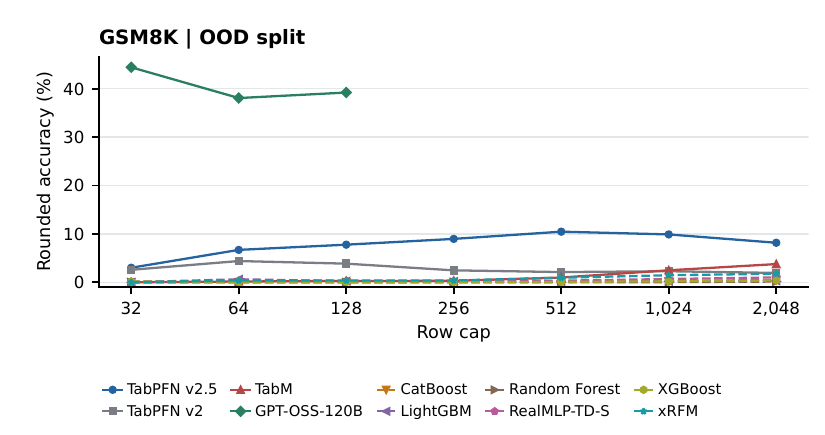}
\caption{\textbf{Every tabular method gains far more from additional rows on the random split than on the OOD split.} GSM8K rounded accuracy across row caps; top: random split, bottom: OOD split. All nine tabular methods use the 100 families at every cap. The GPT-OSS-120B points use the families available at each cap, whose counts appear in Appendix~\ref{app:protocol-coverage}; Figure~\ref{fig:scaling} holds that model's family set fixed across caps. Lines connect the observed caps.}
\label{fig:full-gsm8k}
\end{figure}

\begin{figure}[tp]
\centering
\includegraphics[width=\linewidth]{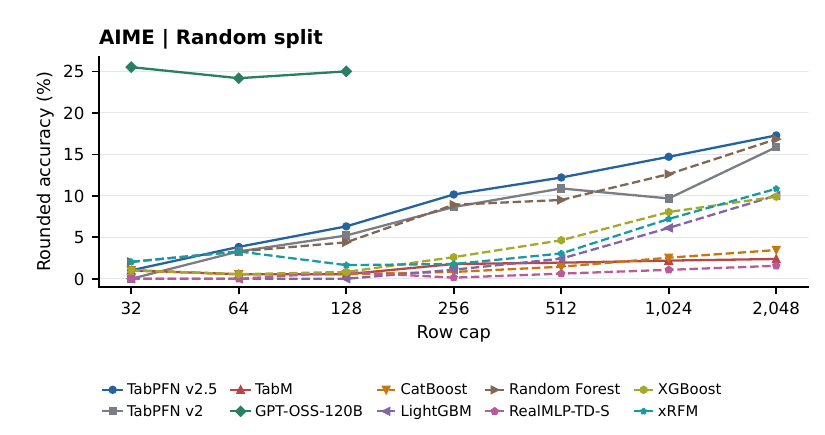}
\includegraphics[width=\linewidth]{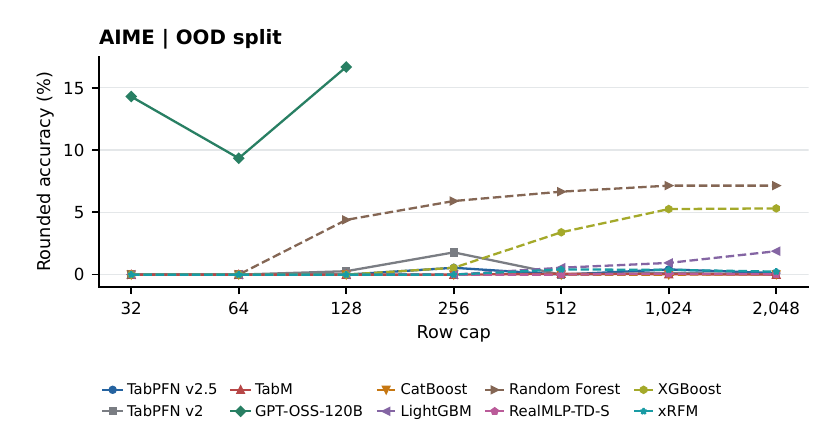}
\caption{\textbf{On all 14 AIME families, OOD accuracy stays low for every method.} Rounded accuracy across row caps; top: random split, bottom: OOD split. These panels include 2024-I-14, whose rounded targets are affected by float32 conversion, and use the GPT-OSS-120B families available at each cap. Figure~\ref{fig:aime} gives the primary analysis on the numerical-resolution-stable families with a fixed ICL family set.}
\label{fig:full-aime}
\end{figure}

Figures~\ref{fig:full-gsm8k} and~\ref{fig:full-aime} show every method across row caps, averaging the families available at each cap. They complement the controlled comparisons of the main text: Section~\ref{sec:llm} compares GPT-OSS-120B and TabPFN v2.5 on the same 87 families, and Figures~\ref{fig:scaling} and~\ref{fig:aime} hold the ICL family set fixed across caps. The full profiles thus describe the whole benchmark, while the main-text figures measure change on a fixed set of families.

\subsection{Tables at Every Row Cap}
The tables in the next two subsections give rounded accuracy at all seven row caps and regression errors at caps 128 to 2,048 for the nine tabular methods. For a family's query set $Q$, let $z$ and $\hat z$ be targets and predictions on the evaluation's standardized target scale. Then
\[
 \operatorname{RMSE}=\sqrt{|Q|^{-1}\!\sum_{j\in Q}(\hat z_j-z_j)^2},
 \qquad
 \operatorname{MAE}=|Q|^{-1}\!\sum_{j\in Q}|\hat z_j-z_j|.
\]
Each entry is the median of a family-level regression metric or the mean of family-level rounded accuracy. Standardized errors describe scale-relative prediction quality, whereas rounded accuracy is computed after returning to the original answer scale. $R^2$ takes the query-set mean as its reference, so a negative value means that the model's squared error on the query set exceeds that of this constant. All entries are computed from the family-level records, and the AIME entries use the same 13 numerical-resolution-stable families as the main analysis. These tables omit GPT-OSS-120B, which was evaluated only at caps 32 to 128.

The complete profiles show that the OOD gap is not a small-data effect. Random-split accuracy typically rises with more rows, whereas the OOD-split profiles remain much flatter, and the regression errors show how much numerical structure survives when the rounded answer is missed. Tables~\ref{tab:full-GSM8K-random}--\ref{tab:full-GSM8K-reg} give the GSM8K profiles and Tables~\ref{tab:full-AIME-random}--\ref{tab:full-AIME-reg} the AIME profiles.
\clearpage
\subsection{GSM8K Accuracy and Error Profiles}
\begin{table}[H]
\centering\small
\setlength{\tabcolsep}{4pt}
\caption{\textbf{More rows raise random-split accuracy for every tabular method.} GSM8K rounded accuracy (\%) at every row cap on the random split. The same 100 families appear in every cell.}
\label{tab:full-GSM8K-random}
\begin{tabular*}{\linewidth}{@{\extracolsep{\fill}}lrrrrrrr@{}}
\toprule
Method & 32 & 64 & 128 & 256 & 512 & 1,024 & 2,048 \\
\midrule
TabPFN v2.5 & 16.0 & 29.2 & 39.5 & 49.3 & 53.6 & 58.4 & 62.0 \\
TabPFN v2 & 16.6 & 24.8 & 25.3 & 30.7 & 36.3 & 41.1 & 46.4 \\
TabM & 3.3 & 5.6 & 9.8 & 12.5 & 18.8 & 25.2 & 31.6 \\
CatBoost & 2.0 & 2.8 & 4.9 & 8.7 & 14.4 & 21.6 & 29.6 \\
XGBoost & 1.6 & 3.5 & 5.2 & 7.5 & 11.3 & 17.5 & 24.0 \\
LightGBM & 2.3 & 2.5 & 5.2 & 7.3 & 12.4 & 17.4 & 23.6 \\
Random Forest & 2.3 & 2.4 & 4.6 & 5.5 & 8.2 & 12.3 & 17.9 \\
RealMLP-TD-S & 2.0 & 3.4 & 4.0 & 5.4 & 9.2 & 14.1 & 19.3 \\
xRFM & 2.7 & 3.9 & 7.0 & 10.9 & 18.8 & 28.2 & 36.8 \\
\bottomrule
\end{tabular*}
\end{table}

\begin{table}[H]
\centering\small
\setlength{\tabcolsep}{4pt}
\caption{\textbf{On the OOD split, no tabular method exceeds 10.5\% at any row cap.} GSM8K rounded accuracy (\%) at every row cap on the OOD split. The same 100 families appear in every cell.}
\label{tab:full-GSM8K-ood}
\begin{tabular*}{\linewidth}{@{\extracolsep{\fill}}lrrrrrrr@{}}
\toprule
Method & 32 & 64 & 128 & 256 & 512 & 1,024 & 2,048 \\
\midrule
TabPFN v2.5 & 3.0 & 6.7 & 7.8 & 9.0 & 10.5 & 9.9 & 8.2 \\
TabPFN v2 & 2.6 & 4.4 & 3.8 & 2.5 & 2.1 & 2.3 & 1.9 \\
TabM & 0.0 & 0.2 & 0.3 & 0.3 & 1.0 & 2.5 & 3.8 \\
CatBoost & 0.0 & 0.0 & 0.0 & 0.1 & 0.1 & 0.2 & 0.5 \\
XGBoost & 0.1 & 0.1 & 0.0 & 0.0 & 0.1 & 0.2 & 0.4 \\
LightGBM & 0.0 & 0.6 & 0.2 & 0.2 & 0.1 & 0.2 & 0.3 \\
Random Forest & 0.0 & 0.0 & 0.0 & 0.0 & 0.0 & 0.0 & 0.1 \\
RealMLP-TD-S & 0.1 & 0.2 & 0.3 & 0.2 & 0.3 & 0.7 & 1.0 \\
xRFM & 0.0 & 0.4 & 0.4 & 0.4 & 0.9 & 1.5 & 1.8 \\
\bottomrule
\end{tabular*}
\end{table}

\clearpage\begingroup\small\setlength{\tabcolsep}{3pt}
\begin{longtable}{llrrrrrr}
\caption{\textbf{On the OOD split, only TabPFN v2.5 and, from 512 rows, TabM have a positive median $R^2$.} GSM8K regression errors at row caps 128 to 2,048. Each entry is the median of family-level scores; RMSE and MAE use training-standardized targets.}\label{tab:full-GSM8K-reg}\\
\toprule
& & \multicolumn{3}{c}{Random} & \multicolumn{3}{c}{OOD} \\ Model & Cap & $R^2$ & RMSE & MAE & $R^2$ & RMSE & MAE \\\midrule
\endfirsthead
\toprule
& & \multicolumn{3}{c}{Random} & \multicolumn{3}{c}{OOD} \\ Model & Cap & $R^2$ & RMSE & MAE & $R^2$ & RMSE & MAE \\\midrule
\endhead
\bottomrule\endfoot
TabPFN v2.5 & 128 & 0.999 & 0.022 & 0.014 & 0.759 & 0.572 & 0.444 \\
 & 256 & 1.000 & 0.010 & 0.006 & 0.756 & 0.460 & 0.374 \\
 & 512 & 1.000 & 0.007 & 0.004 & 0.750 & 0.442 & 0.327 \\
 & 1,024 & 1.000 & 0.007 & 0.004 & 0.593 & 0.525 & 0.335 \\
 & 2,048 & 1.000 & 0.005 & 0.003 & 0.284 & 0.656 & 0.454 \\
TabPFN v2 & 128 & 0.989 & 0.087 & 0.062 & -1.340 & 1.667 & 1.333 \\
 & 256 & 0.995 & 0.069 & 0.032 & -1.303 & 1.549 & 1.224 \\
 & 512 & 0.998 & 0.041 & 0.022 & -1.406 & 1.610 & 1.161 \\
 & 1,024 & 0.999 & 0.033 & 0.016 & -1.418 & 1.564 & 1.193 \\
 & 2,048 & 0.999 & 0.025 & 0.010 & -1.384 & 1.498 & 1.095 \\
TabM & 128 & 0.957 & 0.200 & 0.136 & -1.184 & 1.408 & 1.218 \\
 & 256 & 0.985 & 0.118 & 0.084 & -0.467 & 1.227 & 1.067 \\
 & 512 & 0.994 & 0.070 & 0.046 & 0.027 & 0.959 & 0.808 \\
 & 1,024 & 0.998 & 0.046 & 0.030 & 0.354 & 0.808 & 0.665 \\
 & 2,048 & 0.999 & 0.032 & 0.020 & 0.455 & 0.707 & 0.567 \\
CatBoost & 128 & 0.813 & 0.380 & 0.285 & -5.219 & 2.420 & 2.268 \\
 & 256 & 0.942 & 0.233 & 0.155 & -4.000 & 2.191 & 2.006 \\
 & 512 & 0.979 & 0.135 & 0.088 & -2.807 & 1.948 & 1.748 \\
 & 1,024 & 0.994 & 0.072 & 0.047 & -2.013 & 1.753 & 1.511 \\
 & 2,048 & 0.998 & 0.045 & 0.025 & -1.460 & 1.599 & 1.328 \\
XGBoost & 128 & 0.823 & 0.401 & 0.297 & -4.062 & 2.146 & 1.972 \\
 & 256 & 0.888 & 0.309 & 0.210 & -3.124 & 2.037 & 1.905 \\
 & 512 & 0.961 & 0.196 & 0.135 & -2.501 & 1.922 & 1.734 \\
 & 1,024 & 0.982 & 0.133 & 0.090 & -1.955 & 1.692 & 1.528 \\
 & 2,048 & 0.993 & 0.086 & 0.058 & -1.480 & 1.547 & 1.343 \\
LightGBM & 128 & 0.813 & 0.377 & 0.280 & -3.353 & 2.001 & 1.857 \\
 & 256 & 0.928 & 0.254 & 0.178 & -2.355 & 1.856 & 1.614 \\
 & 512 & 0.969 & 0.170 & 0.110 & -1.884 & 1.718 & 1.503 \\
 & 1,024 & 0.985 & 0.123 & 0.079 & -1.683 & 1.586 & 1.386 \\
 & 2,048 & 0.994 & 0.079 & 0.055 & -1.606 & 1.555 & 1.328 \\
Random Forest & 128 & 0.795 & 0.435 & 0.327 & -4.296 & 2.311 & 2.109 \\
 & 256 & 0.876 & 0.336 & 0.230 & -3.646 & 2.140 & 1.989 \\
 & 512 & 0.935 & 0.249 & 0.185 & -3.351 & 2.078 & 1.911 \\
 & 1,024 & 0.961 & 0.200 & 0.144 & -2.813 & 1.928 & 1.752 \\
 & 2,048 & 0.978 & 0.146 & 0.103 & -2.656 & 1.866 & 1.664 \\
RealMLP-TD-S & 128 & 0.768 & 0.461 & 0.316 & -4.348 & 2.172 & 2.006 \\
 & 256 & 0.890 & 0.315 & 0.224 & -2.811 & 2.003 & 1.799 \\
 & 512 & 0.947 & 0.225 & 0.152 & -1.996 & 1.851 & 1.626 \\
 & 1,024 & 0.980 & 0.140 & 0.093 & -1.685 & 1.623 & 1.386 \\
 & 2,048 & 0.993 & 0.080 & 0.053 & -1.499 & 1.516 & 1.256 \\
xRFM & 128 & 0.898 & 0.320 & 0.203 & -2.985 & 1.938 & 1.742 \\
 & 256 & 0.965 & 0.181 & 0.124 & -2.649 & 1.911 & 1.625 \\
 & 512 & 0.987 & 0.118 & 0.073 & -2.338 & 1.744 & 1.420 \\
 & 1,024 & 0.994 & 0.075 & 0.045 & -1.984 & 1.603 & 1.271 \\
 & 2,048 & 0.998 & 0.041 & 0.026 & -1.716 & 1.538 & 1.172 \\
\end{longtable}\endgroup
\clearpage
\subsection{AIME Accuracy and Error Profiles}
\begin{table}[H]
\centering\small
\setlength{\tabcolsep}{4pt}
\caption{\textbf{On AIME, random-split accuracy peaks at 18.6\%.} AIME rounded accuracy (\%) at every row cap on the random split. The same 13 numerical-resolution-stable families appear in every cell.}
\label{tab:full-AIME-random}
\begin{tabular*}{\linewidth}{@{\extracolsep{\fill}}lrrrrrrr@{}}
\toprule
Method & 32 & 64 & 128 & 256 & 512 & 1,024 & 2,048 \\
\midrule
TabPFN v2.5 & 1.1 & 4.1 & 6.8 & 10.9 & 13.1 & 15.8 & 18.6 \\
TabPFN v2 & 0.0 & 3.6 & 5.6 & 9.3 & 11.7 & 10.4 & 17.1 \\
TabM & 1.1 & 0.6 & 0.6 & 1.9 & 2.1 & 2.4 & 2.6 \\
CatBoost & 1.1 & 0.6 & 0.6 & 0.9 & 1.6 & 2.7 & 3.7 \\
XGBoost & 1.1 & 0.6 & 0.9 & 2.8 & 5.0 & 8.7 & 10.7 \\
LightGBM & 0.0 & 0.0 & 0.0 & 1.2 & 2.6 & 6.6 & 10.8 \\
Random Forest & 2.2 & 3.6 & 4.7 & 9.6 & 10.2 & 13.6 & 18.1 \\
RealMLP-TD-S & 0.0 & 0.0 & 0.9 & 0.1 & 0.7 & 1.2 & 1.7 \\
xRFM & 2.2 & 3.6 & 1.8 & 1.9 & 3.3 & 7.8 & 11.7 \\
\bottomrule
\end{tabular*}
\end{table}

\begin{table}[H]
\centering\small
\setlength{\tabcolsep}{4pt}
\caption{\textbf{On AIME, no tabular method exceeds 7.7\% OOD accuracy at any row cap.} AIME rounded accuracy (\%) at every row cap on the OOD split. The same 13 numerical-resolution-stable families appear in every cell.}
\label{tab:full-AIME-ood}
\begin{tabular*}{\linewidth}{@{\extracolsep{\fill}}lrrrrrrr@{}}
\toprule
Method & 32 & 64 & 128 & 256 & 512 & 1,024 & 2,048 \\
\midrule
TabPFN v2.5 & 0.0 & 0.0 & 0.0 & 0.6 & 0.0 & 0.5 & 0.1 \\
TabPFN v2 & 0.0 & 0.0 & 0.3 & 1.9 & 0.0 & 0.2 & 0.0 \\
TabM & 0.0 & 0.0 & 0.0 & 0.0 & 0.1 & 0.0 & 0.0 \\
CatBoost & 0.0 & 0.0 & 0.0 & 0.0 & 0.0 & 0.0 & 0.0 \\
XGBoost & 0.0 & 0.0 & 0.0 & 0.6 & 3.7 & 5.7 & 5.7 \\
LightGBM & 0.0 & 0.0 & 0.0 & 0.0 & 0.6 & 1.0 & 2.0 \\
Random Forest & 0.0 & 0.0 & 4.7 & 6.4 & 7.2 & 7.7 & 7.7 \\
RealMLP-TD-S & 0.0 & 0.0 & 0.0 & 0.0 & 0.0 & 0.1 & 0.0 \\
xRFM & 0.0 & 0.0 & 0.0 & 0.0 & 0.4 & 0.4 & 0.3 \\
\bottomrule
\end{tabular*}
\end{table}

\clearpage\begingroup\small\setlength{\tabcolsep}{3pt}
\begin{longtable}{llrrrrrr}
\caption{\textbf{On AIME, every tabular method has a negative median OOD $R^2$ at every row cap.} AIME regression errors at row caps 128 to 2,048. Each entry is the median of family-level scores; RMSE and MAE use training-standardized targets.}\label{tab:full-AIME-reg}\\
\toprule
& & \multicolumn{3}{c}{Random} & \multicolumn{3}{c}{OOD} \\ Model & Cap & $R^2$ & RMSE & MAE & $R^2$ & RMSE & MAE \\\midrule
\endfirsthead
\toprule
& & \multicolumn{3}{c}{Random} & \multicolumn{3}{c}{OOD} \\ Model & Cap & $R^2$ & RMSE & MAE & $R^2$ & RMSE & MAE \\\midrule
\endhead
\bottomrule\endfoot
TabPFN v2.5 & 128 & 0.950 & 0.221 & 0.111 & -1.429 & 3.769 & 3.087 \\
 & 256 & 0.969 & 0.218 & 0.081 & -0.753 & 2.909 & 1.984 \\
 & 512 & 0.994 & 0.077 & 0.045 & -0.833 & 2.538 & 1.712 \\
 & 1,024 & 0.993 & 0.060 & 0.026 & -0.773 & 2.519 & 1.673 \\
 & 2,048 & 0.938 & 0.243 & 0.035 & -1.399 & 6.440 & 4.403 \\
TabPFN v2 & 128 & 0.856 & 0.349 & 0.222 & -1.656 & 7.234 & 4.958 \\
 & 256 & 0.844 & 0.321 & 0.134 & -1.316 & 6.476 & 5.356 \\
 & 512 & 0.906 & 0.263 & 0.187 & -1.647 & 6.364 & 4.894 \\
 & 1,024 & 0.899 & 0.299 & 0.177 & -1.302 & 6.550 & 5.159 \\
 & 2,048 & 0.984 & 0.124 & 0.080 & -1.378 & 5.530 & 4.012 \\
TabM & 128 & 0.768 & 0.462 & 0.319 & -1.145 & 6.462 & 5.567 \\
 & 256 & 0.880 & 0.286 & 0.147 & -0.890 & 5.508 & 4.723 \\
 & 512 & 0.962 & 0.160 & 0.118 & -0.705 & 4.068 & 3.024 \\
 & 1,024 & 0.982 & 0.129 & 0.093 & -0.198 & 4.296 & 2.947 \\
 & 2,048 & 0.993 & 0.087 & 0.058 & -0.409 & 3.359 & 2.277 \\
CatBoost & 128 & 0.597 & 0.504 & 0.331 & -2.974 & 7.354 & 6.497 \\
 & 256 & 0.815 & 0.355 & 0.270 & -2.758 & 6.492 & 5.819 \\
 & 512 & 0.905 & 0.264 & 0.182 & -1.736 & 6.146 & 4.898 \\
 & 1,024 & 0.951 & 0.208 & 0.124 & -1.222 & 5.658 & 4.344 \\
 & 2,048 & 0.983 & 0.131 & 0.089 & -1.261 & 4.899 & 3.554 \\
XGBoost & 128 & 0.613 & 0.564 & 0.359 & -2.618 & 7.162 & 6.148 \\
 & 256 & 0.622 & 0.494 & 0.299 & -2.662 & 6.400 & 5.712 \\
 & 512 & 0.864 & 0.317 & 0.228 & -2.772 & 6.075 & 4.896 \\
 & 1,024 & 0.935 & 0.240 & 0.153 & -1.250 & 5.729 & 4.522 \\
 & 2,048 & 0.957 & 0.206 & 0.130 & -1.602 & 5.148 & 3.964 \\
LightGBM & 128 & 0.535 & 0.637 & 0.463 & -2.432 & 6.976 & 5.955 \\
 & 256 & 0.624 & 0.499 & 0.376 & -1.889 & 6.204 & 5.200 \\
 & 512 & 0.903 & 0.267 & 0.187 & -1.377 & 5.728 & 4.522 \\
 & 1,024 & 0.938 & 0.235 & 0.143 & -1.215 & 5.563 & 4.324 \\
 & 2,048 & 0.965 & 0.186 & 0.113 & -1.286 & 5.202 & 3.948 \\
Random Forest & 128 & 0.613 & 0.510 & 0.382 & -2.709 & 7.251 & 6.233 \\
 & 256 & 0.475 & 0.521 & 0.343 & -2.676 & 6.468 & 5.789 \\
 & 512 & 0.808 & 0.377 & 0.276 & -1.830 & 6.251 & 5.092 \\
 & 1,024 & 0.900 & 0.299 & 0.212 & -1.651 & 6.389 & 5.141 \\
 & 2,048 & 0.902 & 0.312 & 0.206 & -1.644 & 6.207 & 4.997 \\
RealMLP-TD-S & 128 & 0.474 & 0.583 & 0.385 & -2.222 & 6.759 & 5.799 \\
 & 256 & 0.636 & 0.481 & 0.312 & -2.389 & 6.625 & 5.797 \\
 & 512 & 0.762 & 0.402 & 0.284 & -1.653 & 6.052 & 4.919 \\
 & 1,024 & 0.934 & 0.252 & 0.190 & -1.196 & 6.087 & 4.699 \\
 & 2,048 & 0.963 & 0.198 & 0.139 & -1.319 & 5.557 & 4.120 \\
xRFM & 128 & 0.547 & 0.507 & 0.335 & -2.375 & 7.351 & 6.496 \\
 & 256 & 0.641 & 0.494 & 0.207 & -2.316 & 6.586 & 5.671 \\
 & 512 & 0.898 & 0.264 & 0.218 & -2.203 & 6.133 & 4.628 \\
 & 1,024 & 0.957 & 0.202 & 0.149 & -1.224 & 6.026 & 4.400 \\
 & 2,048 & 0.979 & 0.148 & 0.110 & -1.467 & 5.996 & 4.306 \\
\end{longtable}\endgroup

\end{document}